\documentclass[10pt,letterpaper]{article}
\usepackage[top=0.85in,left=1.0in,footskip=0.75in,marginparwidth=2in]{geometry}
\usepackage{mathrsfs}
\usepackage{amsthm}
\usepackage{amsmath,amssymb,amsfonts}
\usepackage{bm}
\usepackage[utf8]{inputenc}
\usepackage{cite}
\usepackage{nameref,hyperref}
\usepackage{microtype}
\DisableLigatures[f]{encoding = *, family = * }

\raggedright
\usepackage{changepage}
\usepackage[aboveskip=1pt,labelfont=bf,labelsep=period,singlelinecheck=off]{caption}

\usepackage{color}
\definecolor{Gray}{gray}{.25}
\usepackage{graphicx}
\usepackage{sidecap}

\usepackage{wrapfig}
\usepackage[pscoord]{eso-pic}
\usepackage[fulladjust]{marginnote}
\usepackage{graphicx}
\usepackage[section]{algorithm}
\usepackage{algorithmic}
\usepackage{threeparttable}
\usepackage{bm}
\usepackage{color}
\usepackage{ragged2e}
\usepackage{subfigure}
\usepackage{mathrsfs}
\usepackage{enumitem}
\usepackage{array}
\usepackage{url}
\include{multicol}
\newcommand{\PreserveBackslash}[1]{\let\temp=\\#1\let\\=\temp}
\newcolumntype{C}[1]{>{\PreserveBackslash\centering}p{#1}}
\newcolumntype{R}[1]{>{\PreserveBackslash\raggedleft}p{#1}}
\newcolumntype{L}[1]{>{\PreserveBackslash\raggedright}p{#1}}

\renewcommand{\raggedright}{\leftskip=0pt \rightskip=0pt plus 0cm}

\makeatletter
\renewcommand{\@biblabel}[1]{\quad#1.}
\makeatother

\usepackage{lastpage,fancyhdr,graphicx}
\usepackage{epstopdf}
\fancyheadoffset[L]{2.25in}
\fancyfootoffset[L]{2.25in}

\reversemarginpar

\begin{document}
\vspace*{0.35in}

\begin{flushleft}
{\Large{\textbf{Dynamic Bicycle Dispatching of Dockless Public Bicycle-sharing Systems using Multi-objective Reinforcement Learning}}}
\newline
\\
Jianguo~Chen\textsuperscript{1},
Kenli~Li \textsuperscript{1},
Keqin Li\textsuperscript{1,2},
Philip S. Yu\textsuperscript{3},
Zeng Zeng\textsuperscript{4}
\\
\bigskip
$^{1}$ College of Computer Science and Electronic Engineering, Hunan University, Changsha, Hunan, 410082, China.
\\
$^{2}$ Department of Computer Science, State University of New York, New Paltz, NY, 12561, USA.
\\
$^{3}$ Department of Computer Science, University of Illinois at Chicago, Chicago, IL, 60607, USA.
\\
$^{4}$ Institute for Infocomm Research, Agency for Science Technology and Research (A*STAR), 138632, Singapore.
\bigskip
\\
* Correspinding author: Kenli Li (lkl@hnu.edu.cn).

\end{flushleft}

\section*{Abstract}
As a new generation of Public Bicycle-sharing Systems (PBS), the dockless PBS (DL-PBS) is an important application of cyber-physical systems and intelligent transportation.
How to use AI to provide efficient bicycle dispatching solutions based on dynamic bicycle rental demand is an essential issue for DL-PBS.
In this paper, we propose a dynamic bicycle dispatching algorithm based on multi-objective reinforcement learning (MORL-BD) to provide the optimal bicycle dispatching solution for DL-PBS.
We model the DL-PBS system from the perspective of CPS and use deep learning to predict the layout of bicycle parking spots and the dynamic demand of bicycle dispatching.
We define the multi-route bicycle dispatching problem as a multi-objective optimization problem by considering the optimization objectives of dispatching costs, dispatch truck's initial load, workload balance among the trucks, and the dynamic balance of bicycle supply and demand.
On this basis, the collaborative multi-route bicycle dispatching problem among multiple dispatch trucks is modeled as a multi-agent MORL model.
All dispatch paths between parking spots are defined as state spaces, and the reciprocal of dispatching costs is defined as a reward.
Each dispatch truck is equipped with an agent to learn the optimal dispatch path in the dynamic DL-PBS network.
We create an elite list to store the Pareto optimal solutions of bicycle dispatch paths found in each action, and finally get the Pareto frontier.
Experimental results on the actual DL-PBS systems show that compared with existing methods, MORL-BD can find a  higher quality Pareto frontier with less execution time.

\section{Introduction}
Benefiting from the advantages of zero-emissions, flexibility, and convenience, the Public Bicycle-sharing System (PBS) has become an essential part of urban transportation \cite{wu2020challenges, li2019citywide, ma2019data}.
PBS has significant advantages in short-distance travel, and can be easily combined with public transportation such as buses and subways to solve the problem of ``first mile/last mile'' \cite{zhang2016last}.
In addition to the traditional Station/Dock-based PBS (SD-PBS), many enterprises provide a novel type of Dockless PBS (DL-PBS) system \cite{yoshida2019practical, yang2019mobility}.
The DL-PBS system has the following characteristics: flexible parking spots (also known as parking spots, service points, and bicycle stations), and dynamic changes in the number of bicycles at each parking spot \cite{li2018dynamic}.
Compared with SD-PBS, the significant difference of DL-PBS is that each parking spot (instead of a fixed bicycle station) not only has a flexible location, but also no fixed parking piles/docks \cite{sun2018sharing}.
Users can download a mobile application to quickly identify nearby DL-PBS parking spots via the bicycles' GPS information, and then scan the Quick-Response (QR) code on the target bicycle to unlock and use it \cite{chen2020tii}.
After the trip, users can drop off the bicycle at any parking spots near the destination.
It is very convenient for the public and becomes increasingly popular in many countries.

During the operation of DL-PBS systems, several problems are reported, such as difficulty in picking up or dropping off bicycles, unreasonable station distribution, equipment failures, and serious imbalance between supply and demand \cite{sun2018sharing, yang2019mobility}.
Without the intervention of bicycle dispatching, bicycle parking spots will randomly change during operation, and many destroyed bicycles cannot found in time \cite{pan2019deep, yi2019rebalancing}.
For suppliers, the layout of bicycle parking spots and bicycle dispatching plans are essential to operating profit, bicycle utilization, and user satisfaction.
The optimal layout of bicycle parking spots can maximize the bicycle supply and demand balance by deploying minimum parking spots \cite{yang2019estimating}.
In addition, the optimal bicycle dispatching solution requires the smallest dispatching costs, including the shortest dispatch paths, the least number of dispatched bicycles, and the minimum dispatching time.
Most existing research on public bicycle dispatching focuses on the SD-PBS systems with fixed bicycle stations and fixed parking spots \cite{ren2020rebalancing, li2019citywide}.
However, these methods do not  make full use of large-scale historical riding trajectory records to discover potential riding rules and dynamic rental requirements.

As a new generation of PBS, DL-PBS is an important application of the Internet of Things (IoT), cyber-physical systems (CPS), and Artificial Intelligence (AI) in the field of intelligent transportation \cite{yan2019top, ma2019data}.
In DL-PBS, massive smart bicycles equipped with GPS-based sensors are connected to a complex DL-PBS network, and operate on different spatial and temporal scales through real-time bicycle riding behaviors, exhibiting a variety of distinct behavioral patterns.
In DL-PBS, there is uncertainty in bicycle parking spots, dynamic rental behavior, and the number of bicycles required at each parking spot \cite{chen2020tii, li2019citywide}.
In the process of bicycle dispatching, multiple intelligent dispatch trucks can interact with distributed and moving bicycle sensors in real time, perceive the dynamic supply and demand of parking spots, and dynamically adjust dispatch routes, thereby improving the optimality of dispatch schemes \cite{ghosh2017dynamic}.
Therefore, how to use Machine Learning (ML) and AI approaches to provide efficient bicycle dispatching solutions and meet the dynamic bicycle rental demand is an essential issue for DL-PBS.

\begin{figure}[!ht]
\centering
\includegraphics[width=4.6in]{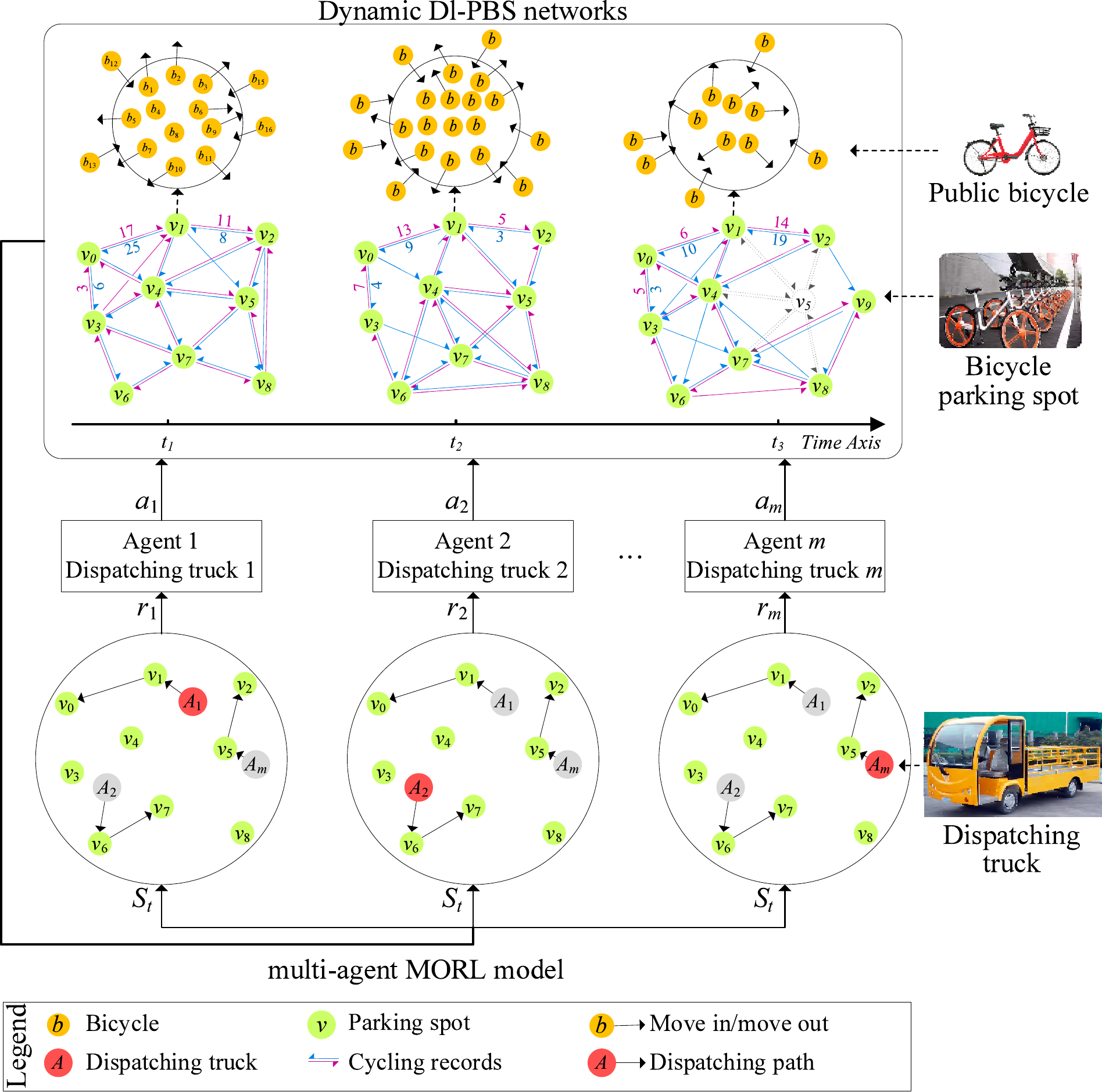}
 \caption{Workflow of the proposed DL-PBS dynamic bicycle dispatching (MORL-BD) algorithm based on multi-objective reinforcement learning.
 Massive public bicycles with intelligent sensors and the corresponding large-scale cycling trajectory records form a series of spatiotemporal DL-PBS networks, and further get a graph sequence model.
 In the MORL-DB algorithm, multiple agents represent multiple dispatch trucks, find the optimal dispatch routes in a cooperative manner, and jointly achieve the optimization goal.}
 \label{fig001}
\end{figure}

In this paper, we propose a Bicycle Dispatching algorithm based on Multi-Objective Reinforcement Learning (MORL-BD) for the DL-PBS system.
The multi-route bicycle dispatching problem is defined as a multi-objective optimization problem.
A multi-agent and MORL algorithm is used to search for candidate Pareto optimal solutions to achieve the expected optimization goals.
Each dispatch truck can automatically detect the bicycles' trajectory and real-time inventory of each parking spot through wireless sensors, and dynamically adjust its dispatch route.
An example of the workflow of the proposed MORL-BD algorithm is shown in Fig. \ref{fig001}.

The contributions of this paper are summarized as follows:

\begin{itemize}
\item We establish a graph sequence model of DL-PBS bicycle stations based on large-scale spatiotemporal datasets, which can richly express the movement behavior of DL-PBS network in different space-time spaces.
    Then, the Gated Graph Neural Network (GGNN) model is used to predict the bicycle station layout and bicycle dispatching demand in the next time period.
\item We define the multi-route bicycle dispatching problem as a multi-objective optimization problem, and well consider four optimization goals, including dispatching costs, the initial load of dispatch trucks, the workload balance between dispatch trucks, and the supply and demand balance of all bicycle stations.
\item We propose a bicycle dispatching (MORL-BD) algorithm based on multi-agent and multi-objective reinforcement learning.
    In MORL-BD, each dispatch truck is equipped with an agent to perceive and interact with the dynamic DL-PBS environment and learn the optimal dispatch path.
    We create an elite list to store the Pareto optimal solution found in each action, and finally find the Pareto frontier.
\item We conduct extensive comparative experiments on actual DL-PBS systems to compare the MORL-BD algorithm with the multi-objective genetic algorithm, evolutionary algorithm, and particle swarm optimization algorithm.
    Experimental results show that MORL-BD can find more high-quality Pareto frontier and the optimal bicycle dispatching plan with less execution time.
\end{itemize}

The rest of the paper is structured as follows.
Section 2 summarizes the related work.
Section 3 describes the DL-PBS system model, the dynamic prediction of bicycle station layout, and the bicycle dispatching problem of public bicycles.
Section 4 introduces reinforcement learning and the proposed MORL-BD algorithm.
Section 5 provides comparison experiments to evaluate the performance of the MORL-BD algorithm.
Finally, Section 6 gives the conclusion.

\section{Related Work}

Accurate prediction of bicycle rental demand is an important prerequisite for public bicycle dispatching and PBS system rebalancing \cite{tang2018bikeshare, mao2019novel}.
In \cite{mimura2019bike}, Mimura \emph{et al}. proposed a time-series generation model to predict the number of bicycle transfers per hour and the bicycle rental demand.
In \cite{li2019citywide}, Li \emph{et al}. introduced an adaptive transition-constrained clustering algorithm to classify public bicycle stations.
They used a similarity-based Gaussian regression to predict the location of different proportions of stations and the bicycle demand of each station.
Considering the relationship between the location of bicycle stations and transport stations, Tang \emph{et al}. carried out a method for adjusting the location and scale of PBS pools in a bike-and-ride multi-modal transport system \cite{tang2018bikeshare}.
In \cite{ghosh2017dynamic}, Ghosh \emph{et al}. discussed the congestion or starvation of bicycle stations caused by unpredictable bicycle rental activities.
They offered an optimization method to predict bicycle riding routes and expected rental demand.
The existing studies are mainly aimed at traditional SD-PBS systems, including bicycle station planning and bicycle-rental demand prediction.
However, there is limited research conducted on emerging DL-PBS systems.

Imbalanced bicycle supply is one of the major problems facing by PBS systems \cite{zhao2019study, hu2014optimal, liu2018vehicle}.
In \cite{zhao2019study}, Zhao \emph{et al}. studied the dispatching and management of DL-PBS systems and established a semi-open dispatching model based on fuzzy time windows.
In \cite{hu2014optimal}, Hu \emph{et al}. built a mathematical model of the location between bicycle stations and dispatch centers, and carried out the optimal dispatching route with the minimized operating cost, passenger travel cost, and dispatching cost.
Different from SD-PBS systems, the bicycle dispatching of DL-PBS systems is more complex due to the uncertainty of the location of bicycle stations and the requirements of bicycles.
In \cite{liu2018vehicle}, Liu \emph{et al}. focused on the DL-PBS bicycle dispatching and divided the DL-PBS networks by the K-means clustering algorithm.
They built a bicycle dispatching model based on a rolling horizon dispatching algorithm, which can effectively guide bicycle redistribution between bicycle stations.
Although there have been some studies on bicycle dispatching and bicycle supply balance, most of them treat the problem as a static problem with single-objective or multi-objective optimization problems.
For the DL-PBS system, due to dynamic bicycle rental demand and dynamic bicycle parking spots, it is difficult to perform bicycle dispatching and maintain the balance of bicycle supply among different bicycle drop-off stations.

Reinforcement learning (RL) methods were introduced in the existing literature to achieve bicycle rebalance without human intervention \cite{li2018dynamic, chen2018personalized, duan2019optimizing, pan2019deep}.
In \cite{li2018dynamic}, Li \emph{et al}. developed a spatiotemporal RL model of bicycle layout, which can reposition bicycle stations and minimize customer losses.
The spatiotemporal RL model is applied to each station cluster to learn the corresponding reposition strategy.
In \cite{chen2018personalized}, Chen \emph{et al}. used the Q-learning algorithm to formulate the personalized bicycle travel plan, and used a dynamic and flexible position insertion method to automatically adjust the trips.
Duan \emph{et al}. discussed the impact of bicycle underflow and overflow on the PBS service and urban traffic congestion \cite{duan2019optimizing}.
In \cite{pan2019deep}, Pan \emph{et al}. constructed a deep RL framework to motivate users to rebalance public bicycles.
They modeled the bicycle rebalancing problem as a Markov decision process, and used the deep deterministic strategy gradient method to capture the spatial and temporal dependence of bicycle stations.
Although these methods provided a variety of incentive mechanisms, due to factors such as user riding purpose and participation, these methods only played a limited role in practical applications without the intervention of manual dispatching, which makes it difficult to rebalance the system.

In current research of multi-objective optimization (MOO) problems and multi-objective reinforcement learning (MORL), the standard concept of optimality is replaced by Pareto optimality \cite{parisi2016multi, hu2012data, ruiz2017temporal}.
In \cite{parisi2016multi}, Parisi \emph{et al}. formulated an RL strategy gradient method to learn Pareto boundary in multi-objective Markov decision problems, where the continuous approximation of Pareto boundary is generated for each gradient climb operation.
In \cite{ruiz2017temporal}, Ruiz \emph{et al}. discussed a MORL method that uses non-convex Pareto boundaries to generates deterministic non-dominated strategies in multi-objective Markov decision problems.
To solve the effect of MORL on the optimal strategy under different preference conditions, Yang \emph{et al}. adopted a MORL algorithm with linear preference \cite{yang2019generalized}.
Even though different RL and MORL approaches have been proposed for MOO problems and applications, the existing MORL methods have parameter configuration problems and scalarization limitations.
In addition, in our work, we define the multi-route bicycle dispatching problem as a multi-objective optimization problem by considering multiple conflicting optimization objectives.
In this case, the standard MORL methods may face limitations and inefficiently find the Pareto frontier of the proposed problem.

Different from the existing work of DS-PDS networks, we focus on the dynamic prediction of bicycle station layout and bicycle dispatching requirements in actual DL-PBS networks.
We model the DL-PBS network from the perspective of cyber-physical systems and construct a graph sequence model of bicycle parking spots based on historical spatiotemporal cycling and dispatching trajectory records.
In addition, different from traditional single-truck-based dispatching, we defined the problem as a multi-route bicycle dispatching problem.
Moreover, we apply the MORL algorithm to the multi-route bicycle dispatching problem and use multiple agents to collaborate and interact with the dynamic DL-PBS environments, which can efficiently find the Pareto frontier between multiple conflicting objectives.

\section{System Model and Problem Formulation}
In this section, we will describe the DL-PBS system model and use a clustering algorithm and a deep learning model to predict the dynamic layout of bicycle parking spots and dispatching requirements.
In addition, the bicycle dispatching problem is formulated as a multi-objective optimization problem by considering four conflicting objectives.

\subsection{DL-PBS System Model}
\begin{figure}[!ht]
\centering
\includegraphics[width=3.4in]{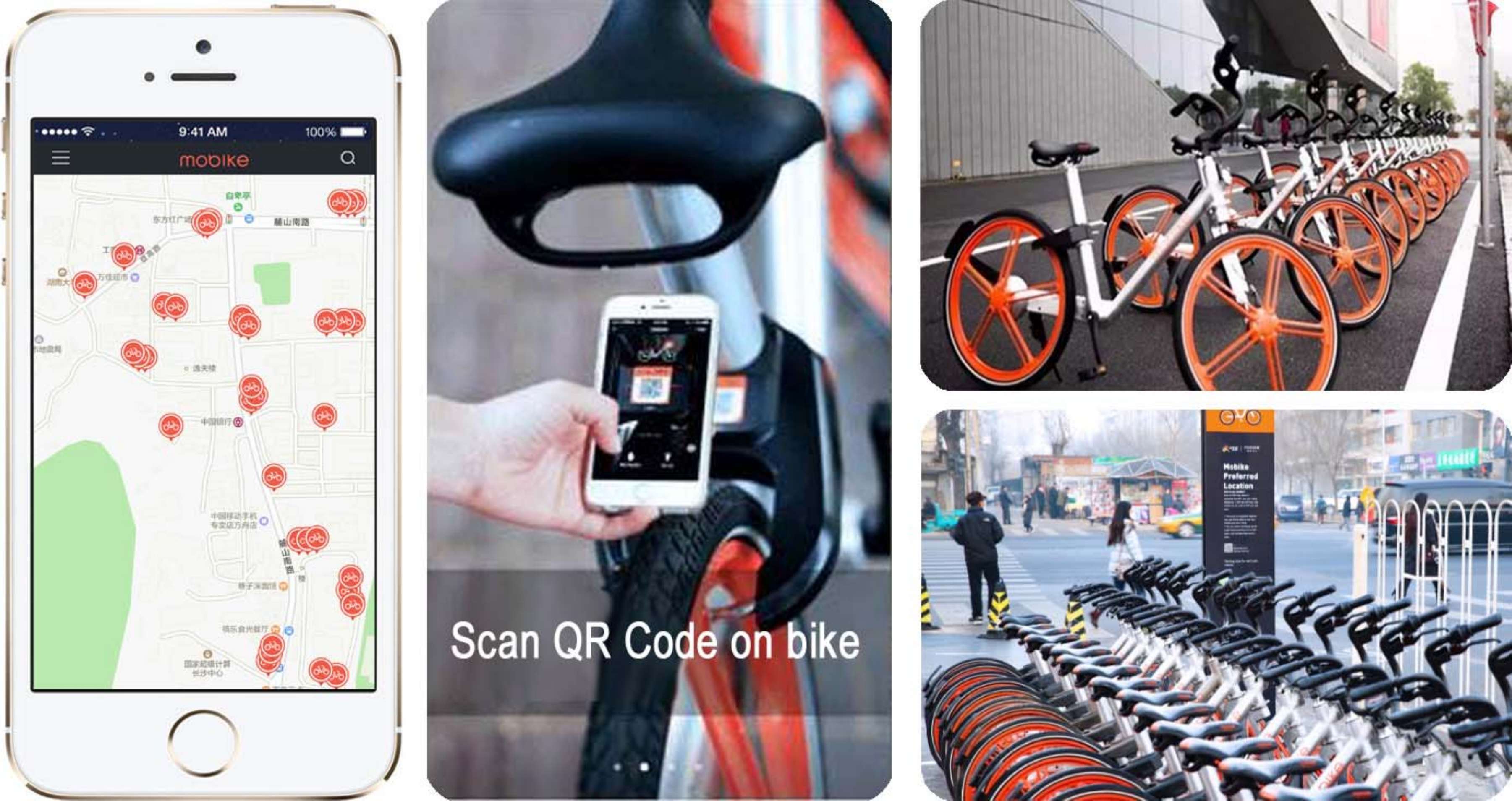}
 \caption{Example of the main components of the DL-PBS system, where the left sub-figure is a mobile application, the middle is a bicycle with a lock module based on Quick Response (QR) code, and the right is a bicycle parking spot.}
 \label{fig002}
\end{figure}

Different from the traditional SD-PBS systems, multiple DL-PBS suppliers deploy their own DL-PBS systems in each city.
They are allowed to place a large number of public bicycles in all permitted parking areas, such as road-sides, parks, entrances to communities, and shopping centers.
A certain number of bicycles parked in the same location will naturally form a bicycle parking spot, also known as drop-off/pick-up position, a self-service point, and a bicycle station.
Each parking spot has no strict boundaries and a limitation to the number of bicycles.
In this way, the bicycle station layout of DL-PBS is more flexible than that of SD-PBS, and the number of bicycles provided at each parking spot is not limited by parking piles or docks.
Therefore, the location and scale of bicycle spots can be adjusted dynamically according to the demand of cycling, so as to achieve high flexibility.
In this work, a DL-PBS system include five main components, including public bicycles, bicycle parking spots, dispatching centers, a mobile application, and a bicycle dispatching and management system.
An example of the main components of the DL-PBS system is shown in Fig. \ref{fig002}.

(1) Public bicycles.
Each DL-PBS supplier provides a large number of public bicycles with unique appearance.
Each bicycle is equipped with a global positioning system (GPS) module to record its position in real time.
In addition, it is also equipped with a lock module based on Quick Response (QR) code.

(2) Dockless bicycle parking spots.
DL-PBS suppliers deploy public bicycles in permitted parking areas of the city, such as roadsides, parks, entrances to communities, and shopping centers.
A dense group of bicycles forms a temporary dockless bicycle parking spot.
Note that there are no strict restrictions on the location and size of parking spots, as well as no fixed parking piles or docks at each parking spot.
When all bicycles at a parking spot are removed, the parking spot will disappear automatically.
On the contrary, when several bicycles are parked in the same location, a bicycle parking spot will be formed naturally.
In this way, DL-PBS suppliers can easily move bicycles and parking spot at low cost.

(3) Bicycle dispatching centers.
Multiple bicycle dispatching centers are deployed in different areas of each city.
Each dispatching center has multiple dispatch trucks, which are responsible for dispatching bicycles in a limited area.
Note that the location of each dispatching center is fixed.

(4) Mobile application (App).
Users can download a mobile application from the DL-PBS website or scan the QR-code on any public bicycle.
The App provides functions including bicycle GPS positioning, QR code scanning, unlocking, payment, and bicycle tracking.
Namely, users can find the nearest bicycles via the App, scan the QR code on a bicycle to unlock it, pay the rent after the travel.
They also can log in the App to track their historical riding records.

(5) Bicycle dispatching and management system.
The bicycle dispatching and management system is responsible for maintaining and managing the basic information of all bicycles, parking spots, and dispatching centers of the entire DL-PBS system.
It also contains functions such as bicycle positioning, trajectory tracking, parking spot prediction, and dispatching plan recommendation.

\subsection{Layout Prediction of Bicycle Parking Spots}

In this work, historical bicycle GPS and trajectory records of the DL-PBS system are used to construct the corresponding graph model of bicycle parking spots.
According to the administrative region of the city and the time periods, historical records are divided into a series of spatiotemporal data subsets.
Each subset represents the bicycle GPS information and corresponding cycling records of an administrative area (e.g., city, administrative district, or county) in a certain time period (e.g., one day or one week).
In our previous work \cite{chen2020tii}, we used the Domain Adaptive Density Clustering (DADC) algorithm to cluster bicycle parking spots, and then constructed a weighted directed graph model based on the clustering results.

We establish a weighted digraph model $G = (V, ~E)$ for the DL-PBS network in each spatiotemporal subset, where the set of vertices $V$ represents the bicycle parking spots, and the set of edges $E$ represents the cycling trajectories between parking spots.
Each vertex $v_{i} \in V$ has three attribute values ($\mu_{i}$, $\psi_{i}$, $\varphi_{i}$), where $\mu_{i}$ indicates the number of bicycles parked at the spot, and $\psi_{i}$ and $\varphi_{i}$ indicate the latitude and longitude of $v_{i}$.
Each edge $e_{ij} \in E$ has two attribute values $(d_{ij}, w_{ij})$, which indicate the actual distance between spots $v_{i}$ and $v_{j}$ and the corresponding number of cycling records.
Note that $d_{ij} = d_{ji}$, but $w_{ij} \neq w_{ji}$.

\begin{figure}[!ht]
\setlength{\abovecaptionskip}{0pt}
\setlength{\belowcaptionskip}{0pt}
\centering
\includegraphics[width=5.0in]{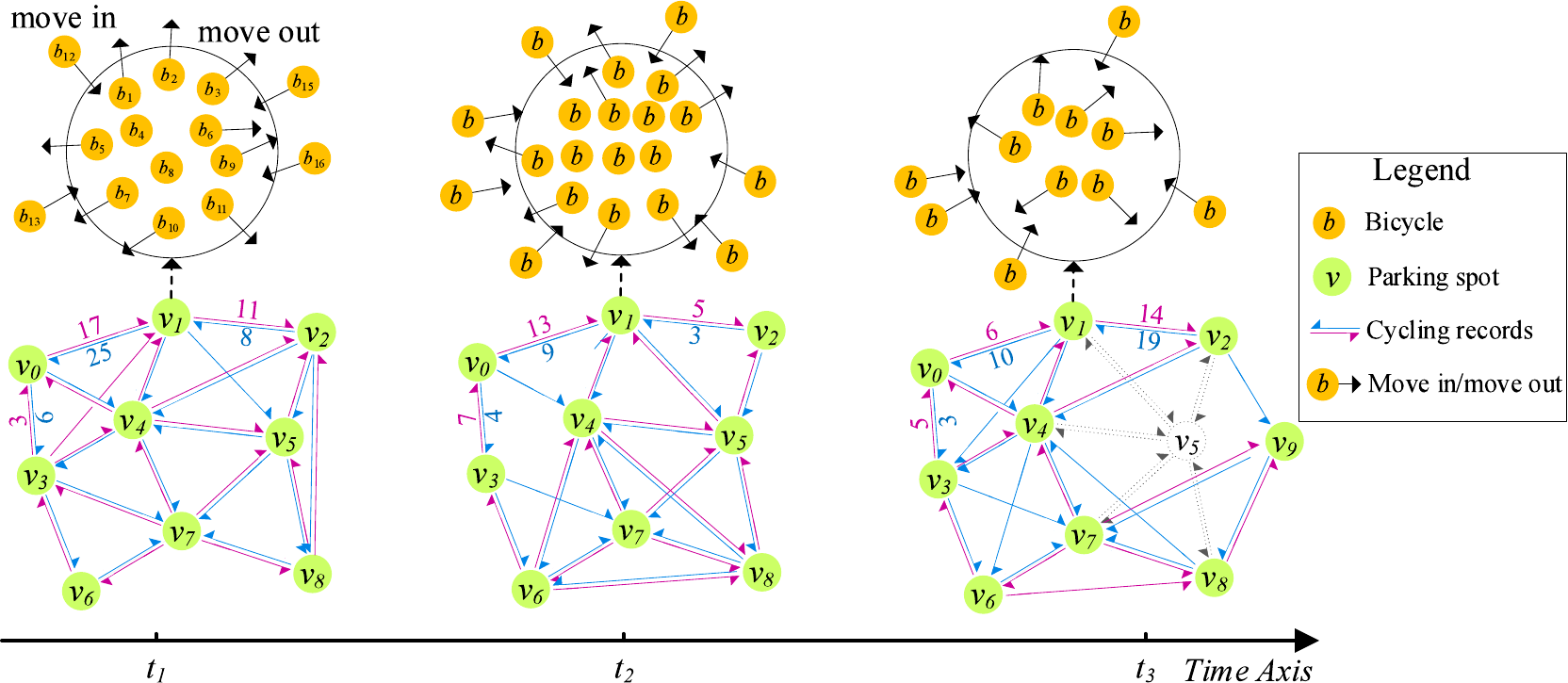}
\caption{Example of the graph sequence model of DL-PBS. Each graph represents the graph model of the DL-PBS network in a spatiotemporal subset. At each vertex (a parking spot), we can observe the return (move in) and rental (move out) behavior of bicycles, which leads to vertex update.}
\label{fig003}
\end{figure}

We further study the dynamic behavior of bicycle rental and return and the update of parking spots.
With the rental and return of bicycles, the number of bicycles available at each parking spot dynamically changes.
In addition, new spots may appear and some current spots may disappear.
We perform bicycle parking spot clustering on each spatiotemporal subset to construct the corresponding graph model, and then combine graph models from multiple time periods to construct a graph sequence model.
An example of the graph sequence model of DL-PBS is shown in Fig. \ref{fig003}.

Let $GS = \{G_{1}, \ldots, G_{t}\}$ be a DL-PBS graph sequence model of a certain spatial subset.
According to the latitude and longitude coordinates of each vertex, we establish a connection between the vertices of different graph models.
In other words, vertices with the same or similar latitude and longitude coordinate values are regarded as the same vertices in the graph sequence model.
Then, we can calculate the update of each vertex across time periods.
Considering the matching errors of vertices caused by the imprecise GPS information, we introduce a fusion factor $\epsilon$ for vertex connection.
Namely, for a vertex $v_i$ in graph $G_t$ and a vertex $v_j$ in graph $G_{t+1}$, if $|\varphi_i-\varphi_j| \leq \epsilon$ and $|\psi_i-\psi_j |\leq \epsilon$, vertices $v_i$ and $v_j$ are treated as the same vertex.
In practical applications, the value of $\epsilon$ is manually set based on experience.
In this work, the effective threshold is set as $\epsilon=3$ (m).

Assuming that $\mu_{i,t}$ and $\mu_{i,t+1}$ are the number of bicycles at $v_{i}$ in $G_{t}$ and $G_{t+1}$, respectively.
The updating of each vertex $v_{i}$ between two adjacent graph models $G_{t}$ and $G_{t+1}$ can be divided into five cases.
\begin{enumerate}
\item The number of bicycles remains the same at $v_{i}$.
      If $v_{i}$ exists in both $G_{t}$ and $G_{t+1}$, and $\mu_{i,t+1} =\mu_{i,t}$, it means that the same number of bicycles at $v_{i}$ is maintained between $t$ and $t+1$.
       Namely, the number of bicycles arriving at $v_{i}$ is equal to the number of bicycles leaving $v_{i}$.
\item  The number of bicycles at $v_{i}$ decreases.
       If $v_{i}$ exists in both $G_{t}$ and $G_{t+1}$, and $\mu_{i,t+1} < \mu_{i,t}$, it means that the number of bicycles at $v_{i}$ decreases.
       Namely, the number of bicycles arriving at $v_{i}$ is less than the number of bicycles leaving $v_{i}$.
\item  The number of bicycles at $v_{i}$ increases.
       If $v_{i}$ exists in both $G_{t}$ and $G_{t+1}$, and $\mu_{i,t+1} > \mu_{i,t}$, it means that the number of bicycles at $v_{i}$ decreases between $t$ and $t+1$.
       Namely, the number of bicycles arriving at $v_{i}$ is more than the number of bicycles leaving $v_{i}$.
\item  New parking spot emerges.
       If $v_{i}$ does not appear in $G_{t}$ but in $G_{t+1}$, it means that $v_{i}$ is newly generated in the period between $t$ and $t+1$.
       $\mu_{i,t+1}$ is the number of bicycles at $v_{i}$ in $t+1$, which is equal to the number of bicycles arriving in time point $t+1$  minus the number of bicycles departing.
\item  Parking spot disappears.
       If $v_{i}$ only appears in $G_{t}$ but does not appear in $G_{t+1}$, it means that $v_{i}$ disappears between time points $t$ and $t+1$.
\end{enumerate}

\begin{figure}[!ht]
\centering
\includegraphics[width=5.4in]{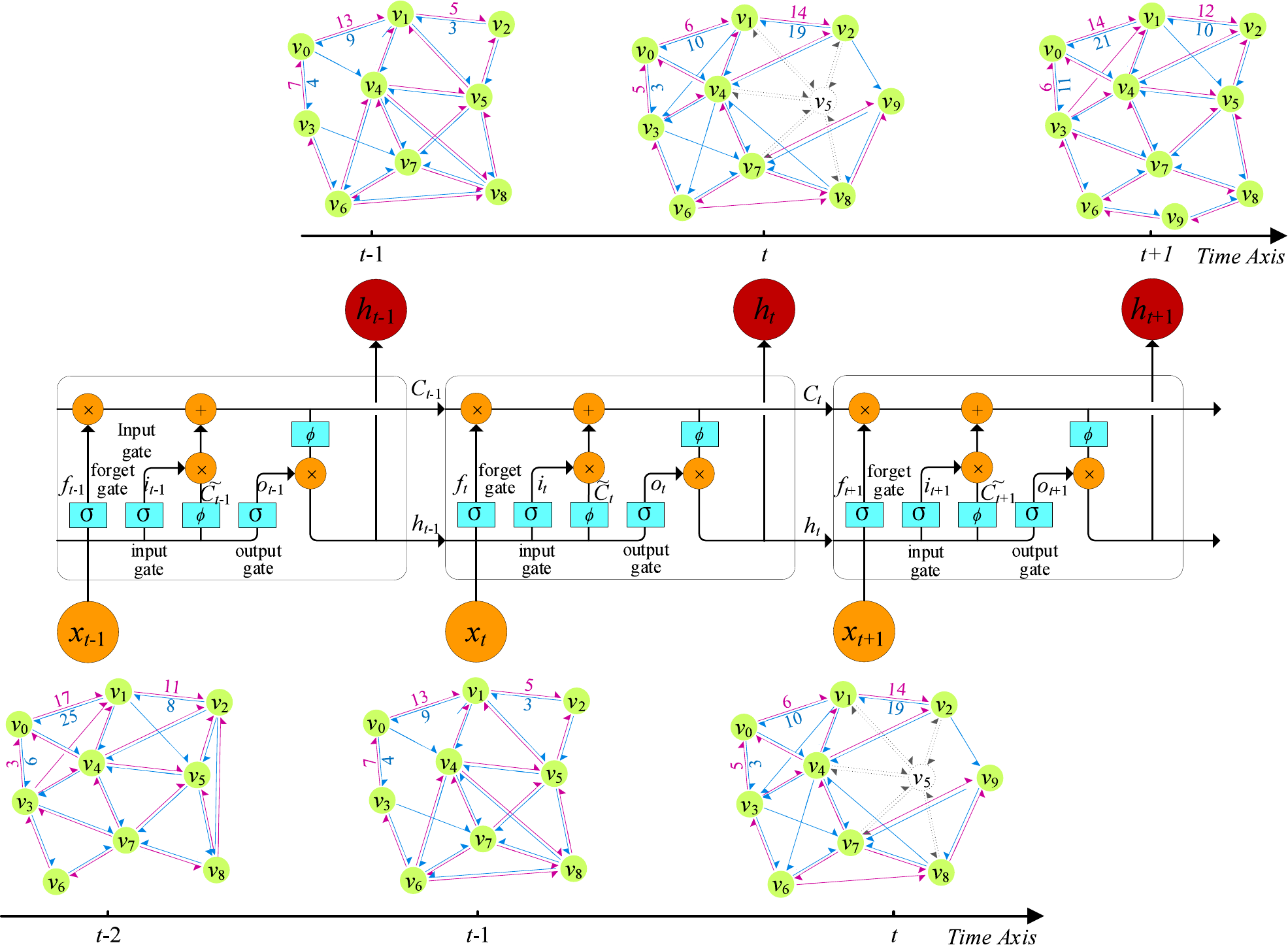}
 \caption{The structure of the GGNN model for bicycle station prediction. The input of GGNN is the constructed graph sequences $\{G_{1}, \ldots, G_{t}\}$ of historical bicycle parking spots, and the output is the predicted graph model $G_{t+1}$ for the next time period.}
 \label{fig004}
\end{figure}

We trained the Gated Graph Neural Network (GGNN) model \cite{gnn08} on the constructed graph sequence models and used the GGNN model to predict the bicycle station layout for the next time period.
For specific implementation details, please refer to our previous work \cite{chen2020tii}.
The structure of the GGNN model for bicycle station prediction is shown in Fig. \ref{fig004}.

As shown in Fig. \ref{fig004}, based on the graph sequence $\{G_{1}, \ldots, G_{t}\}$ of historical bicycle parking spots, we use the GGNN model to predict the graph model $G_{t+1}=(V_{t+1}, E_{t+1})$ for the next time period $t+1$.
In the set of predicted vertices $V_{t+1}$, we obtain the position ($\psi_{i}$, $\varphi_{i}$) of each vertex $v_{i}$ in time period $t+1$ and the number of bicycles $\mu_{i}$ required at $v_{i}$.
At the same time, we also obtain the set of predicted edges $E_{t+1}$ between vertices in $V_{t+1}$.
For each edge $e_{ij} \in E_{t+1}$, we calculate its actual distance $d_{ij}$ and predict the number of cycling records $w_{ij}$  between vertices $v_{i}$ and $v_{j}$.

\subsection{Bicycle Dispatching Requirements}

\begin{figure}[!ht]
\centering
\includegraphics[width=3.4in]{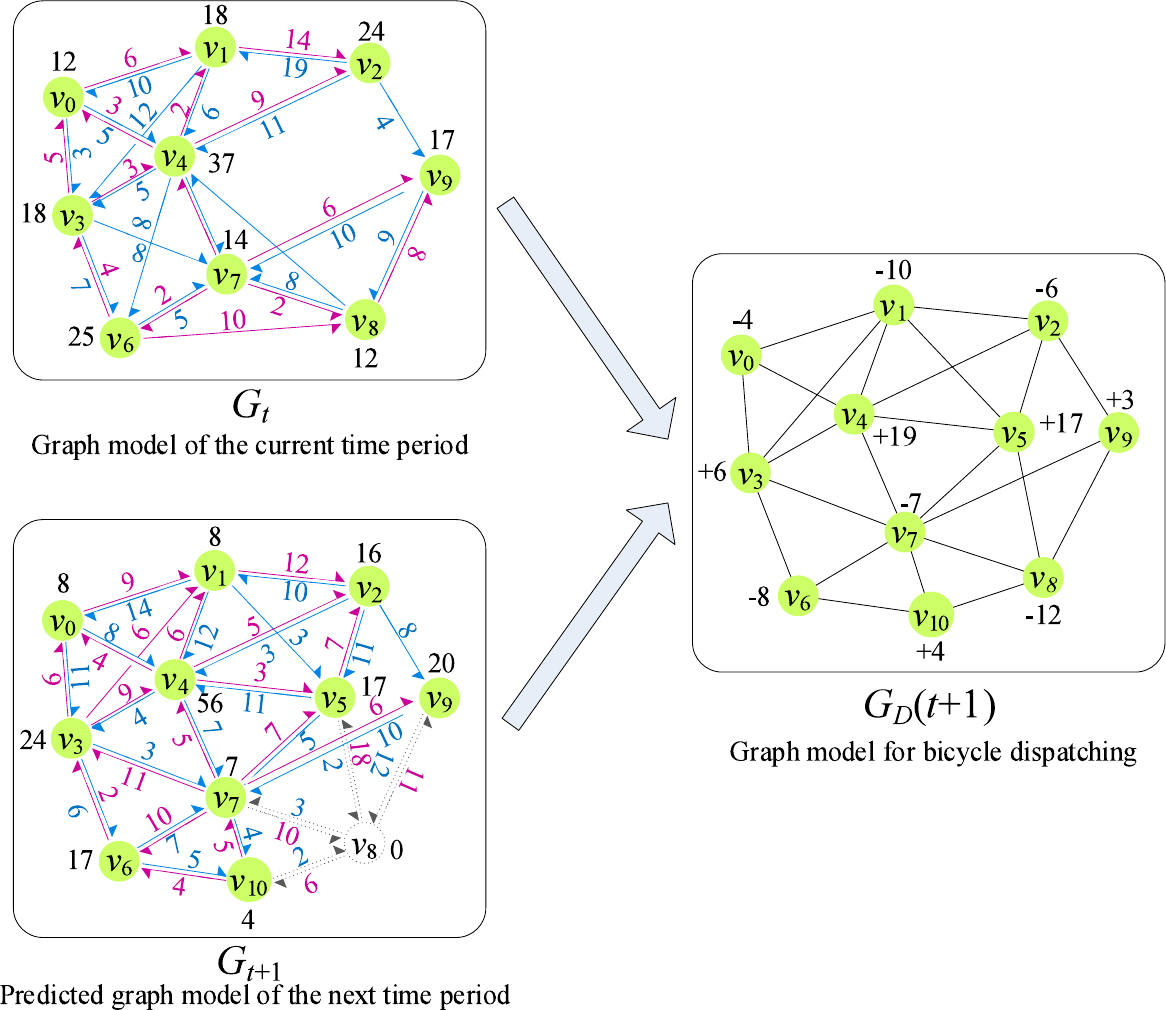}
 \caption{Construction process of bicycle dispatching demand graph model. Based on the graph model of the current time period and the predicted graph model of the next time period, the bicycle dispatching demand graph model in the next time period is constructed by considering the update of vertices.}
 \label{fig005}
\end{figure}

Based on the actual graph model $G_{t}$ of the current time period and the predicted graph model $G_{t+1}$ of the next time period, we can calculate the bicycle dispatching requirements in the next time period.
Then, the corresponding bicycle dispatching demand graph model $G^{D}_{t+1}$ is established.
An example of bicycle dispatching demand calculation is illustrated in Fig. \ref{fig005}.

Given $G_{t}=(V_{t}, E_{t})$ and $G_{t+1}=(V_{t+1}, E_{t+1})$, we establish a dispatching demand graph model $G^{D}_{t+1}= (V^{D}_{t+1}, E^{D}_{t+1})$ of the time period $t+1$, which is an undirected complete graph.
For each vertex $v_{i} \in ( V_{t} \cap V_{t+1})$, if $\mu_{i,t} = \mu_{i,t+1}$, then note it as a stable spot.
The set of all stable spots in  $V_{t}$ and $V_{t+1}$ is defined as $V^{S}_{t,t+1} = \{v_{i}|v_{i} \in V_{t} \cap V_{t+1} ~~\text{and}~~  \mu_{i,t} = \mu_{i,t+1}\}$.
We create a set of vertices $V^{D}_{t+1}$ based on $V_{t}$ and $V_{t+1}$:
\begin{equation}
V^{D}_{t+1} = (V_{t} \cup V_{t+1}) - V^{S}_{t,t+1},
\end{equation}
namely, all vertices in $V_{t}$ and $V_{t+1}$ are firstly included into $V^{D}_{t+1}$, and then all stable vertices in $V^{S}_{t,t+1}$ are removed from $V^{D}_{t+1}$.
We continue to calculate the number of bicycles to be dispatched at each vertex $v_{i}$:
\begin{equation}
\mu^{D}_{i,t+1} = \mu_{i,t+1} - \mu_{i,t}.
\end{equation}
In addition, we create the edges $E^{D}_{t+1}$ based on $E_{t}$ and $E_{t+1}$.
Different from the directed edges in $E_{t}$ and $E_{t+1}$, we define the undirected edges $E^{D}_{t+1}$ for $G^{D}_{t+1}$:
\begin{equation}
E^{D}_{t+1} = E_{t} \cup E_{t+1}.
\end{equation}
Then, for each edge $e_{ij} \in E^{D}_{t+1}$, we remove the opposite edge $e_{ji}$ from $E^{D}_{t+1}$ if $e_{ji} \in E^{D}_{t+1}$.
The detailed steps of constructing the DL-PBS bicycle dispatching demand graph model are described in Algorithm \ref{alg1}.
Assuming that $N$ is the number of vertices (bicycle stations) in the predicted graph model $G_{t+1}$ and $M$ is the number of edges (trajectory routes) in $G_{t+1}$, the time complexity of Algorithm 3.1 is $O(M+N)$.

\begin{algorithm}[!ht]
\caption{Construction of bicycle dispatching demand graph model of DL-PBS}
\label{alg1}
\begin{algorithmic}[1]
\REQUIRE ~\\
    $G_{t}$: the actual graph model at the current time period $t$;\\
    $G_{t+1}$: the predicted graph model for the next time period $t+1$;\\
\ENSURE ~\\
    $G^{D}_{t+1}$: the bicycle dispatch graph model of $t+1$.
\STATE Combine vertices from $V_{t}$ and $V_{t+1}$: $V^{D}_{t+1} \leftarrow V_{t} \cup V_{t+1}$;
\FOR {each vertex $v_{i}$ in $V^{D}_{t+1}$}
\IF {$v_{i} \in V_{t}$ and $v_{i} \in V_{t+1}$ and $\mu_{i,t} = \mu_{i,t+1}$}
\STATE Remove $v_{i}$ from $V^{D}_{t+1}$;
\ELSE
\STATE Calculate the number of bicycles to be dispatched $\mu^{D}_{i,t+1} \leftarrow \mu_{i,t+1} - \mu_{i,t}$;
\ENDIF
\ENDFOR
\STATE Combine edges from $E_{t}$ and $E_{t+1}$: $E^{D}_{t+1} \leftarrow E_{t} \cup E_{t+1}$;
\FOR {each edge $e_{ij}$ in $E^{D}_{t+1}$}
\IF {$e_{ij} \in E^{D}_{t+1}$ and $e_{ji} \in E^{D}_{t+1}$}
\STATE Remove $e_{ji}$ from $E^{D}_{t+1}$;
\ENDIF
\ENDFOR
\STATE Create a bicycle dispatch graph model $G^{D}_{t+1} \leftarrow (V^{D}_{t+1}, E^{D}_{t+1})$;
\RETURN $G^{D}_{t+1}$.
\end{algorithmic}
\end{algorithm}

\subsection{Problem Formulation}
Based on the bicycle dispatching requirements, we define the bicycle dispatching problem in this section.
Generally, there are multiple DL-PBS dispatching centers in each city, each dispatching center has multiple dispatch trucks, and each dispatch truck is responsible for several bicycle parking spots.
Given a bicycle dispatching demand graph model $G^{D}_{t+1}$ and a set of dispatch trucks $K =\{K_{1}, \ldots, K_{M}\}$.
In this way, the bicycle dispatching optimization problem is transformed into a traveling salesman problem with several closed loops, namely, the multi-route bicycle dispatching problem.
Each dispatch truck executes a dispatching loop, which is defined as follows.

\textbf{Definition 1: Dispatching loop.}
The dispatching loop is a directed closed link and is completed by a dispatch truck.
The dispatch truck loads a certain number of bicycles from a dispatching center according to the dispatching plan, completes the bicycle deployment tasks at each parking spot in the dispatching plan, and then returns to the dispatching center.

In each dispatching loop, the number of parking spots, the route of dispatching, and the number of initially loaded bicycles on the trucks are dynamically determined according to the corresponding dispatching plan.

\textbf{Definition 2: Multi-route bicycle dispatching solution.}
In a DL-PBS system of a spacial subset (e.g., a city), given the bicycle dispatching demand graph model $G^{D}_{t+1}$ and a set of dispatch trucks $K =\{K_{1}, \ldots, K_{M}\}$, each dispatch truck completes a disjoint dispatching loop.
All dispatching loops form a multi-route bicycle dispatching solution.

The model parameters of the multi-route bicycle dispatching problem are defined as follows:
\begin{itemize}
\item $G^{D}_{t+1} = (V^{D}_{t+1}, E^{D}_{t+1})$: the bicycle dispatching demand graph model of time period $t+1$;
\item $K = \{K_{1}, \ldots, K_{M}\}$: all dispatch trucks;
\item $Q$: the maximum carrying capacity of each dispatch truck;
\item $\tau_{d}$: the travel time cost of a dispatch truck within unit distance;
\item $\tau_{u}$: the operation time cost of loading and unloading each bicycle;
\item $\mu^{D}_{i}$: the number of bicycles to be dispatched at spot $v_{i}$;
\item $e_{ij} \in [0,~1]$: the edge between vertices $v_{i}$ and $v_{j}$, if an edge exists between $v_{i}$ and $v_{j}$, then $e_{ij}=1$, otherwise, $e_{ij}=0$;
\item $d_{ij}$: the distance of edge $e_{ij}$;
\item $L = \{L_{1}, \ldots, L_{m}, \ldots, L_{M}\}$: the set of dispatching loops of all dispatch trucks;
\item $L_{m} = \{e_{ij}\}$: the set of edges (dispatch paths) in the $m$-th dispatching loop;
\item $q_{m}$: the number of bicycles currently loaded on the $m$-th dispatch truck;
\item $q^{0}_{m}$: the number of bicycles initially loaded on the $m$-th dispatch truck;
\item $q^{Max}_{m}$: the maximum number of bicycles loaded on the $m$-th dispatch truck.
\end{itemize}

\subsubsection{Optimization Objectives}
In this work, the optimization objectives of the multi-route bicycle dispatching problem include four aspects: minimizing the dispatching costs, minimizing the number of initially loaded bicycles on the trucks, maximizing the workload balance between dispatch trucks, and maximizing of the supply and demand balance of all parking spots.

(1) Minimum dispatching costs.

The bicycle dispatching cost includes the travel cost $C_{\text{travel}}$ of dispatching loops and the time cost $C_{\text{time}}$ of loading and unloading bicycles.
The length of dispatch paths refers to the sum of the length of dispatching loops completed by all dispatch trucks.
Therefore, we define the travel cost of all dispatching loops as:
\begin{equation}
\label{eq04}
C_{\text{travel}} = \sum_{L_{m} \in L}{\sum_{e_{ij} \in L_{m}}}{d_{ij}}.
\end{equation}
The time cost $C_{\text{time}}$ of dispatching loops refers to the travel time of dispatching loops and the operation time of loading and unloading bicycles at each parking spot:
\begin{equation}
C_{\text{time}} = \sum_{L_{m} \in L}{\sum_{e_{ij} \in L_{m}}}{\left(d_{ij} \times \tau_{d} + \mu^{D}_{i} \times \tau_{u}\right)}.
\end{equation}

Since the number of bicycles to be dispatched in $G^{D}_{t+1}$ is fixed, the time cost of loading bicycles in the entire graph is fixed.
In addition, the time cost of travel is proportional to the distance of dispatching loops.
Therefore, it is easy to prove that the time cost of dispatching loops is linearly related to the travel cost, and the minimum value of the travel cost must lead to the minimum time cost.
In the following, we only consider the travel cost as dispatching cost.

(2) Minimum initial load of the dispatch trucks.

\begin{figure}[!ht]
\centering
\includegraphics[width=5.0in]{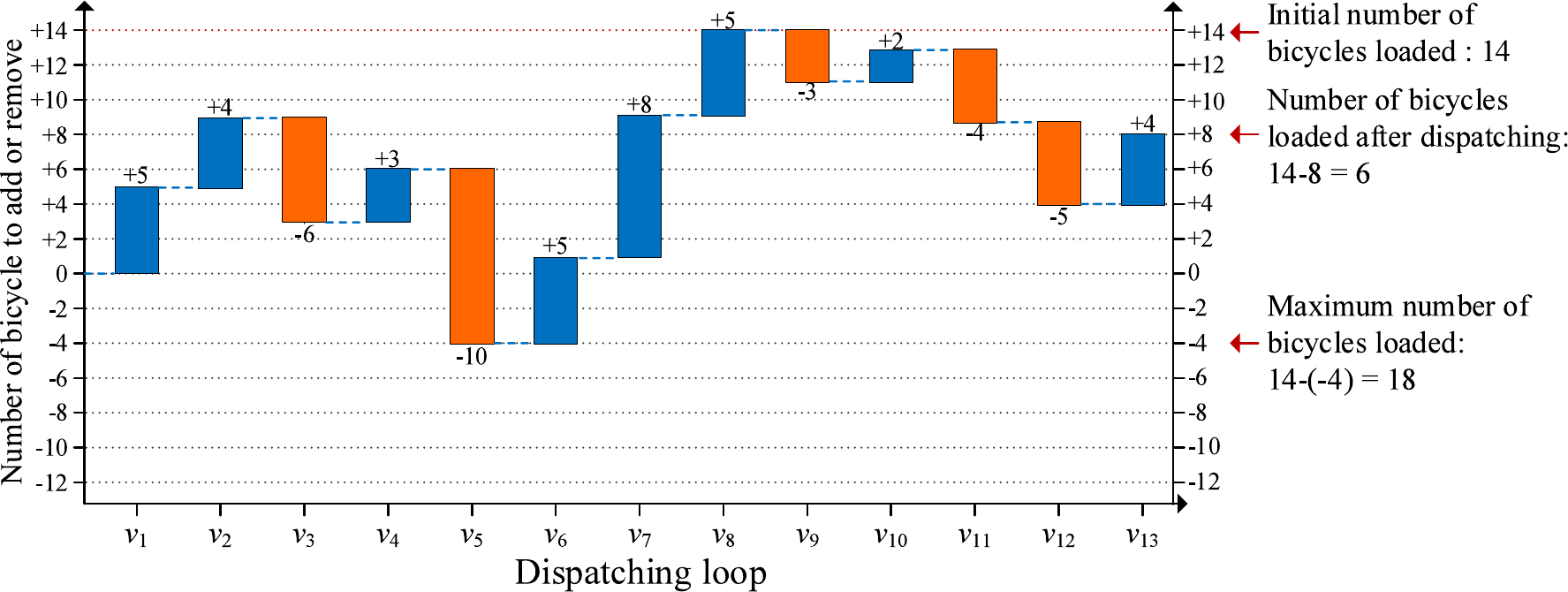}
 \caption{Example of calculation of the number of bicycles initially loaded on the trucks and the maximum number of bicycles loaded. There are 13 vertices (+5, +4, -6, +3, -10, +5, +8, +5, -3, +2, -4, -5, +4) in a dispatching loop, where a positive number indicates that the vertex needs to add bicycles, and a negative number indicates that the vertex needs to remove bicycles. We can calculate the number of bicycles initially loaded is equal to 14 by using Eq. (\ref{eq07}), the maximum number of bicycles loaded is equal to 18 by using Eq. (\ref{eq08}).}
 \label{fig006}
\end{figure}

In the bicycle dispatching process, some parking spots need to add bicycles (unload from the dispatch truck), while others need to remove bicycles (load to the dispatch truck).
In this way, the dispatch trucks do not need to start at full-load in the initial state.
For a certain dispatching loop, we can calculate the number of bicycles to be dispatched at each bicycle spot and obtain the number of bicycles to be initially loaded to the current dispatch truck.
Assuming that $\mu^{D}_{i}$ is the number of bicycles to be dispatched at parking spot $v_{i}$, $q_{m}^{i}$ is the number of bicycles that is required to be loaded or the number of available spaces on the $m$-th dispatch truck, as defined as:
\begin{equation}
\begin{aligned}
q_{m}^{1} &= \mu^{D}_{1}; \\
q_{m}^{2} &= \mu^{D}_{1} + \mu^{D}_{2};\\
& \ldots\\
q_{m}^{i} & =\sum_{v_{i} \in e_{ij}, e_{ij} \in L_{m}}{\mu^{D}_{i}}.\\
\end{aligned}
\end{equation}
The number $q^{0}_{m}$ of bicycles required to be initially loaded on the $m$-th dispatch truck is calculated as:
\begin{equation}
\label{eq07}
q^{0}_{m} =  \max{\left(q_{m}^{1}, ~q_{m}^{2}, ~\ldots,   ~q_{m}^{i}, ~\ldots\right)}.
\end{equation}
where $q_{m}^{i}$ is the number of bicycles currently loaded on the $m$-th dispatch truck.
Note that each dispatching loop $L_{m}$ is a directed closed link and the order of edges (dispatch paths) and vertices (parking spots) in $L_{m}$ is fixed.
Based on $q^{0}_{m}$, we can further calculate the maximum number of bicycles loaded on the $m$-th dispatch truck:
\begin{equation}
\label{eq08}
q^{Max}_{m} = q^{0}_{m} - \min{\left(q_{m}^{1}, ~q_{m}^{2}, ~\ldots,  ~q_{m}^{i}, ~\ldots\right)}.
\end{equation}
Hence, we define the number of bicycles initially loaded on all dispatch trucks as:
\begin{equation}
q^{0} =\sum_{L_{m} \in L}{q^{0}_{m}}.
\end{equation}
Given a dispatching loop with 13 vertices, an example of the calculation of the initial load and maximum load of the current dispatch truck is illustrated in Fig. \ref{fig006}.

(3) Maximum workload balance of the dispatch trucks.

The workload of each dispatch truck is related to the length of the dispatch loop and the number of bicycles dispatched.
Therefore, the dispatching time cost of the dispatching loop is used as the dispatching workload of each dispatch truck.
We use the reciprocal of the standard deviation of dispatching time cost to calculate the workload balance $B_{W}$ between dispatch trucks, as defined as:
\begin{equation}
B_{W} = \sqrt{\left(\frac{1}{M}\sum_{L_{m} \in L}{\left(C_{\text{time},m} - \overline{C_{\text{time}}}\right)}^{2}\right)^{-1}},
\end{equation}
where $M$ is the number of dispatch trucks, $C_{\text{time},m}$ is the dispatching time cost of the $m$-th dispatching loop, and $\overline{C_{\text{time}}}$ is the average value of the dispatching time cost of all dispatching loops.

(4) Maximum balance of the supply and demand of all bicycle spots.

Since the bicycle rental and return situations of each spot changes over time, when a dispatch truck arrives at a target spot, the number of bicycles that need to be dispatched at the spot is not always equal to the number in the dispatch plan.
Therefore, the number of bicycles to be dispatched needs to be adjusted dynamically to meet the expected demand.
For example, when adding bicycles to a spot, if the number of bicycles available on the dispatch truck is less than the number of bicycles required at the target spot, and the difference is less than the threshold value of $\varepsilon_{d}$, we will continue to perform dispatching at the spot with the available number of bicycles on the truck.
Assuming that spot $v_{i}$ requires to add $\mu^{D}_{i}$ bicycles, but there are $q_{m} (0 < \mu^{D}_{i} - q_{m} \leq \varepsilon_{d})$ bicycles loaded on the $m$-th dispatch truck.
In this case, after adding $q_{m}$ bicycles to $v_{i}$, the difference between the supply and demand of $v_{i}$ is:
\begin{equation}
\psi_{i} = q_{m} - \mu^{D}_{i}.
\end{equation}
Conversely, when removing bicycles from a spot, if the available free space on the dispatch truck is less than the number of bicycles to be removed, we only remove the limited number of bicycles from the target spot.
Assuming that spot $v_{i}$ requires to remove $\mu^{D}_{i}$ bicycles, but there are $Q-q_{m} (0 < \mu^{D}_{i} - (Q-q_{m}) \leq \varepsilon_{d})$ free space available on the dispatch truck.
In this case, after removing $q_{m}$ bicycles from $v_{i}$, the difference between the supply and demand of $v_{i}$ is:
\begin{equation}
\psi_{i} = \mu^{D}_{i} - q_{m}.
\end{equation}
Therefore, the balance between the supply and demand in the dispatching loops is the sum of the absolute value of the balance between the supply and demand of all spots in all dispatch loops, which is defined as:
\begin{equation}
\label{eq13}
B_{S} =\sum_{L_{m} \in L}{\sum_{v_{i} \in L_{m}}{\left(\exp{(-|\psi_{i}|)}\right)}}.
\end{equation}

\textbf{Definition 3: Multi-objective Bicycle Dispatching (MBD) problem.}
Based on the above description, the conflicting multi-objective target considered in the multi-route bicycle dispatching (MBD) problem is comprised of four aspects: minimum dispatching costs,
minimum initial load of the dispatch trucks, maximum workload balance of the dispatch trucks, and maximum balance of the supply and demand of all bicycle spots.
The multi-objective bicycle dispatching problem is formalized as follows:
\begin{equation}
\label{eq14}
\begin{aligned}
\text{Minimize:}~~ F_{1} &= \sum_{L_{m} \in L}{\sum_{e_{ij} \in L_{m}}}{d_{ij}},\\
\text{Minimize:}~~ F_{2} &= \sum_{L_{m} \in L}{q^{0}_{m}},\\
\text{Maximize:}~~ F_{3} &= \sqrt{\left(\frac{1}{M}\sum_{L_{m} \in L}{\left(C_{\text{time},m} - \overline{C_{\text{time}}}\right)}^{2}\right)^{-1}},\\
\text{Maximize:}~~ F_{4} &= \sum_{L_{m} \in L}{\sum_{v_{i} \in L_{m}}{\left(\exp{(-|\psi_{i}|)}\right)}},\\
\end{aligned}
\end{equation}
where function $F_{1}$ is to minimize the dispatching costs, function $F_{2}$ is to minimize the number of bicycles initially loaded on all dispatch trucks, function $F_{3}$ is to maximize the balance of the workload between dispatch trucks, and function $F_{4}$ is to maximize the balance of the supply and demand of all bicycle spots.

\subsubsection{Constraints}
The multi-route bicycle dispatching problem is subjected to the following constraints.

(1) In a dispatching solution, there is no overlap between all dispatching loops:
\begin{equation}
L_{1} \cap L_{2} \cap \cdots \cap L_{|L|} = \emptyset.
\end{equation}

(2) In a dispatching solution, each vertex $v_{i}$ can only be visited once, namely:
\begin{equation}
\sum_{j=1, j \neq i}^{N}{e_{ij}} = 1,
\end{equation}
where $N$ is the number of vertices in $V^{D}_{t+1}$.

(3) The number of bicycles currently loaded on each dispatch truck $K_{m}$ does not exceed its maximum carrying capacity $q_{m}$ at any time:
\begin{equation}
q_{m} \leq Q, \forall K_{m} \in K.
\end{equation}

(4) The number $\mu^{D}_{i}$ of bicycles dispatched at any parking spot $v_{i}$, including adding to or removing from the parking spot, must not exceed the maximum carrying capacity of the dispatch truck:
\begin{equation}
\mu^{D}_{i} \leq Q, \forall v_{i} \in V^{D}_{t+1}.
\end{equation}

\subsubsection{Pareto Optimality}

Since there is usually no unique and perfect solutions for multi-objective problems, we use Pareto optimal solutions \cite{ben1980characterization} to find the optimal solutions to the bicycle dispatching problem.
The Pareto optimal solution is applied in the objective function to obtain a set of non-inferior solution vectors, which can form the Pareto frontier in the target space.
We define the Pareto dominance and Pareto optimality of the MBD optimization problem as follows.

\textbf{Definition 4: Pareto dominance of the MBD problem}.
Given the multi-objective optimization function $F = (F_{1}, F_{2}, F_{3}, F_{4})$ in Eq. (\ref{eq14}) of the MDB problem with four objectives, and $\vec{U}=[u_{1}, u_{2}, u_{3}, u_{4}]$ and $\vec{V}=[v_{1}, v_{2}, v_{3}, v_{4}]$ are two candidate values of this function.
If and only if solution $\vec{U}$ is partially less than solution $\vec{V}$, then $\vec{U}$ dominates $\vec{V}$.
That is, for each component $u_{i}$ and $v_{i}$ in $\vec{U}$ and $\vec{V}$, we can obtain that $u_{i} \leq v_{i}$.

\textbf{Definition 5: Pareto optimality of the MBD problem}.
In the MBD problem, given a solution $\vec{U}=[u_{1}, u_{2}, u_{3}, u_{4}]$, if and only if there is no an alternative solution $\vec{V}=[v_{1}, v_{2}, v_{3}, v_{4}]$ that satisfies $\vec{V}$ dominates $\vec{U}$, then, solution $\vec{U}$ is Pareto optimal.

\section{Bicycle Dispatching Based on Multi-objective Reinforcement Learning}
In this section, we will introduce the multi-agent multi-objective reinforcement learning (multi-agent MORL) to find the Pareto optimal solution of the MBD problem.
We will define the state space, available actions, immediate rewards, and learning process in detail.

\subsection{Multi-objective Reinforcement Learning}
The target problem of a single agent RL model is usually described as a Markov Decision Process (MDP), while a multi-agent MORL model is usually described as a Markov game, where the learning task corresponds to a quaternion $E~=~\langle{M}, S,~A,~P,~R\rangle$, where
$M$ is the number of players, namely, agents and dispatch trucks;
$S = \{s_{1}, \ldots, s_{|S|}\}$ denotes the state space of the target problem, which is a joint state of multiple agents;
$A = \{A_{1}, \ldots, A_{M}\}$ denotes the set of actions available in each state;
$P$ denotes the probability of state transition by using various actions in each state;
and $R$ represents the rewards obtained by the actions executed in each state.
For the current state $s$, we can execute a joint action $a =(a_{1}, \ldots, a_{M})$ to transfer to the next potential state $s'$ with the transition probability $p(s'|s, a)$ and a reward $r(s, a)$.

The purpose of RL is to find the optimal strategy $\pi$ to maximizes the expected reward $R$.
The goal can be represented by a $Q$-value, which records the expected reward for each state-action pair:
\begin{equation}
\label{eq19}
Q(s,a) \leftarrow Q(s,a) + \alpha \left[r(s,a)+\gamma \max_{a'}{Q(S',a')} - Q(s,a)\right],
\end{equation}
where $r$ is the reward for reaching $s'$ state, $\gamma$ is a discount factor, and $\alpha$ is the learning rate.

Different from the single-objective RL algorithms, in the multi-objective RL (MORL) algorithm, the target problem is updated to a multi-objective problem to find multiple conflicting optimization goals.
In this way, the reward $r$ in Eq. (\ref{eq19}) is updated to a reward vector $\vec{r}$.

\subsection{MORL-based Bicycle Dispatching (MORL-BD)}
In the MORL-BD algorithm, we provide a bicycle dispatching demand graph model $G^{D}_{t+1}$ and a set of dispatch trucks $K =\{K_{1}, \ldots, K_{M}\}$.
In this section, we will describe the state space, available actions, immediate rewards, and learning process of MORL-BD to find the Pareto optimal solution.

\subsubsection{State Space of MORL-BD}
In the MORL-BD algorithm, the state space represents all bicycle parking spots that can be visited by all dispatch trucks and the number of bicycles needs to be dispatched in each dispatching loop.
Let $S = \{s_{1}, \ldots, s_{|S|}\}$ be the state space of the MORL-BD algorithm, and $|S|$ is the candidate states.
Since the number of parking spots is $N$ and the number of dispatch trucks is $M$, the average number of spots responsible for each dispatch truck is $\lceil N/M \rceil$.
For each state $s_{i} \in S$, the next state $s'$ represents the set of all the next possible spots and the number of bicycles to be dispatched at the next spot.
In the multi-agent MORL model with multiple dispatching loops, each state can be expressed as $s_{i} = \{s_{i,1}, \ldots, s_{i,M}\}$.
$s_{i, m} \in s_{i}$ is the set of all dispatching tasks assigned to the $m$-th dispatch truck $K_{m} \in M$ at state $s_{i}$.
In our implementation, the current value of the decision variable is used to determine the parking spot to be accessed and the number of bicycles to be dispatched in the current state $s_{i}$.

In each decision state, each dispatch truck performs only one dispatching task by using available actions, that is, visiting a bicycle spot and dispatching a certain number of bicycles.
The maximum decision states of the MORL-BD problem depends on the total number of parking spots in the entire dispatching demand graph and the workload of each dispatch truck.
Note that $|S| < (N + 2)$, where $s_{1}$ is the initial state from the dispatching center to the first bicycle parking spot, and $s_{N}$ is the final state from the last bicycle parking spot in a dispatching loop to the dispatching center.

\subsubsection{Action Set}
\begin{figure}[!ht]
\centering
\includegraphics[width=4.0in]{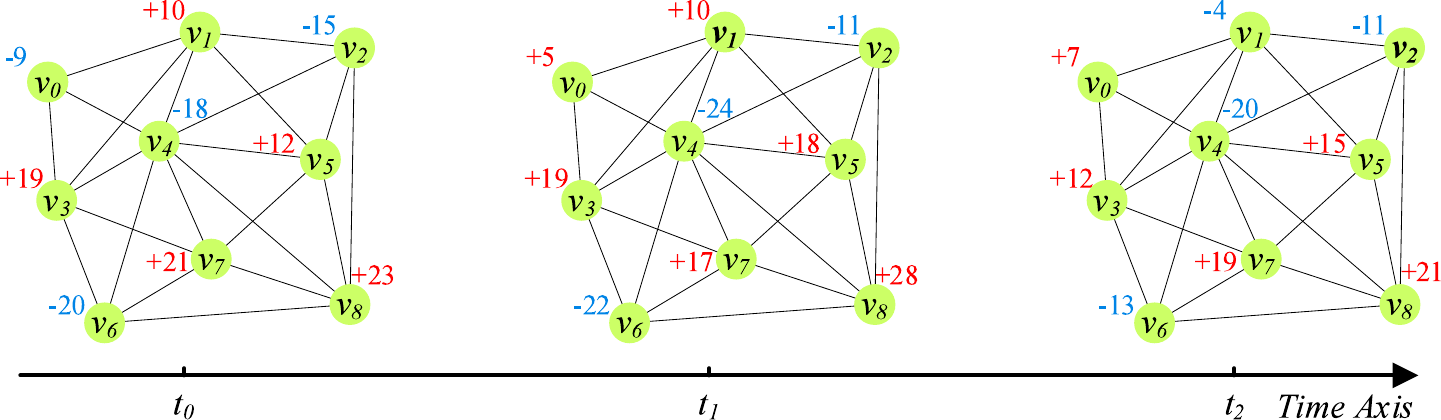}
 \caption{Example of the relationship between dynamic bicycle dispatching requirements and available actions over different time periods. Assuming that the current state is $s=(v_{1},+10)$, the candidate states at $t_{0}$ are $\{s'_{1}=(v_{2},-15), s'_{2}=(v_{5},+12), s'_{3}=(v_{4},-18), s'_{4}=(v_{3},+19)\}$. The candidate states of $s$ at $t_{1}$ are updated to $\{s'_{1}=(v_{2},-11), s'_{2}=(v_{5},+18), s'_{3}=(v_{4},-24), s'_{4}=(v_{3},+19)\}$.}
 \label{fig007}
\end{figure}

The action set of the MORL-BD algorithm refers to the set of all actions that can be executed in each state.
Because the dispatching demand graph is not a fully connected graph, and the number of bicycles to be dispatched at each parking spot is different, the set of actions available to each dispatch truck is different in each state.
In addition, due to the real-time rental and returning behavior of each parking spot in the dispatching process, the actual number of bicycles to be dispatched at each spot may be different from the dispatching plan.
In this case, the actions available in the current state $s_{i}$ are represented as $A_{i} = \{(s'_{1}, a_{1}, p_{1}, r_{1}), \ldots, (s'_{|A_{i}|}, a_{|A_{i}|}, p_{|A_{i}|}, r_{|A_{i}|})\}$.
Each candidate state $s'$ is expressed as $s' = (v_{j}, \mu^{D}_{j})$.
That is, $v_{j}$ is the bicycle parking spot maybe visited by the current dispatch truck, and $\mu^{D}_{j}$ is the number of bicycles to be dispatched  at $v_{j}$.
Fig. \ref{fig007} shows an example of the relationship between dynamic bicycle dispatching requirements and available actions over different time periods.

\subsubsection{Immediate Reward Vector}

In the MORL-BD algorithm, we use an immediate reward vector instead of a single reward for each action.
By performing action $a_{i}$, the immediate reward vector of each state $s_{i}$ is defined as:
\begin{equation}
\label{eq20}
\vec{r}(s_{i},a_{i}) = [\vec{r}^{\,1}(s_{i},a_{i}) , \vec{r}^{\,2}(s_{i},a_{i}) , \vec{r}^{\,3}(s_{i},a_{i}) , \vec{r}^{\,4}(s_{i},a_{i}) ]^{T},
\end{equation}
where $\vec{r}^{\,1}(s_{i},a_{i}) = [r^{1}_{i, 1}, \ldots, r^{1}_{i, M}]$ is the joint rewards of the optimization goal of dispatching costs of all agents (dispatch trucks),
$\vec{r}^{\,2}(s_{i},a_{i}) = [r^{2}_{i, 1}, \ldots, r^{2}_{i, M}]$ is the joint rewards of the optimization goal of the number of bicycles initially loaded on all dispatch trucks,
$\vec{r}^{\,3}(s_{i},a_{i}) = [r^{3}_{i, 1}, \ldots, r^{3}_{i, M}]$ is the joint rewards of the optimization goal of the workload balance between dispatch trucks,
and $\vec{r}^{\,4}(s_{i},a_{i}) = [r^{4}_{i, 1}, \ldots, r^{4}_{i, M}]$ is the joint rewards of the optimization goal of the supply and demand balance of all bicycle spots.
For the current state $s_{i} = (v_{i}, \mu^{D}_{i})$ at decision step $i$ by executing action $a_{i} \in A$, the reward components $r^{1}_{i, m}$ to $r^{4}_{i, m}$ of the $m$-th agent (dispatch truck) are calculated as follows:
\begin{equation}
\begin{aligned}
r^{1}_{i, m} &= \frac{1}{d_{ji}}\sum_{L_{m} \in L}{\sum_{e_{ji} \in Lm}{d_{ji}}},\\
r^{2}_{i, m} &= \frac{1}{q^{0}_{m}}\sum_{L_{m} \in L}{q^{0}_{m}},\\
r^{3}_{i, m} &= \frac{B'_{W}}{B''_{W}},\\
r^{4}_{i, m} &= \frac{B'_{S}}{B''_{S}},\\
\end{aligned}
\end{equation}
where $e_{ji} = (v_{j}, v_{i})$, and $v_{j}$ is the previous bicycle parking spot of $v_{i}$ visited by the same dispatch truck in the previous state.
$B'_{W}$ is the workload balance of the dispatching solution when $v_{i}$ is included in $L_{m}$, while
$B''_{W}$ is the corresponding value when $v_{i}$ is not included in $L_{m}$.
This explanation applies to $B'_{S}$ and $B''_{S}$.

\subsubsection{State Value Evaluation}

\begin{figure}[!ht]
\centering
\includegraphics[width=3.0in]{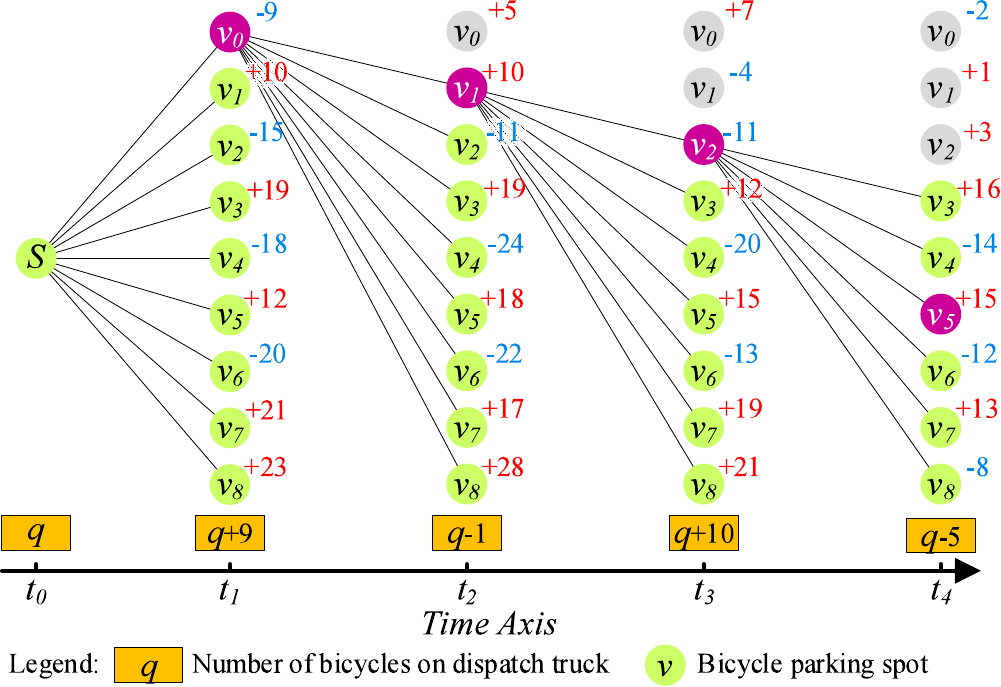}
 \caption{Example of state transition in a bicycle dispatching loop based on the dispatching demand graph model in Fig. \ref{fig007}.}
 \label{fig008}
\end{figure}

In each bicycle dispatching loop $L_{m}$, the immediate reward vector $\vec{r}(s_{i},a_{i})$ is calculated by applying the components $[r_{i,1}, r_{i,2}, r_{i,3}, r_{i,4}]$ in the objective functions in Eq. (\ref{eq13}).
For each dispatch truck, since the current state $s_{i}$ is located at spot $v_{i}$, we update the state value of the current state by calculating:
\begin{equation}
\label{eq22}
\vec{U}(s_{i},a_{i}) \leftarrow \vec{U}(s_{i},a_{i}) + \alpha \left\{\vec{r}(s_{i},a_{i})+\gamma \left[\max_{a' \in A}{\vec{U}(s',a')} - \vec{U}(s_{i},a_{i})\right]\right\},
\end{equation}
where $a'$ is the candidate actions in the state $s_{i}$, and is consists of the collection of all connected parking spots and the number of bicycles to be dispatched on the target spots.
Then, we compare the reward vector with the existing Pareto optimal vectors to decision the candidate actions with the highest immediate rewards.
Once a Pareto optimal solution $\vec{U}=[u_{1}, u_{2}, u_{3}, u_{4}]$ is found, we save it into an elite list and finally form the Pareto frontier.
An example of state transition in a bicycle dispatching loop is shown in Fig. \ref{fig008}.

\begin{figure}[!ht]
\centering
\includegraphics[width=4.5in]{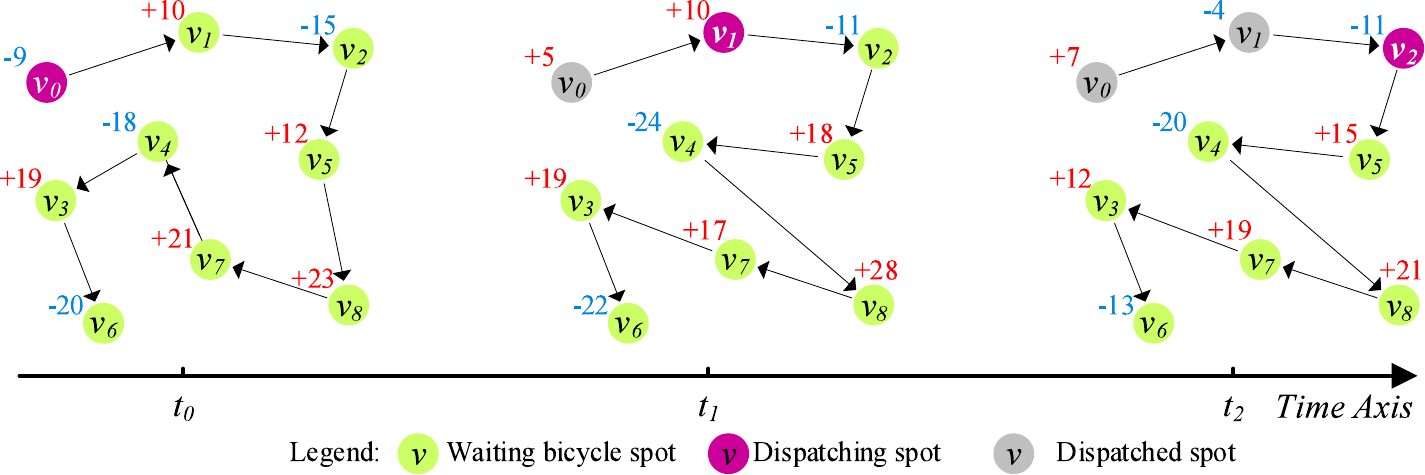}
 \caption{Dynamic update of the dispatching loop based on real-time bicycle rental behavior and state transition in Fig. \ref{fig008}.}
 \label{fig009}
\end{figure}

In Fig. \ref{fig008}, the state transfer is started from the dispatching center to the first bicycle parking spot by using the MORL-BD algorithm.
At each state $s_{i}$, we can generate the candidate actions $A_{i}=\{a_{i}\}$ with all candidate subsequent states.
In addition, we continue to calculate the immediate reward vector $\vec{r}(s',a')$ for each action selection $(s', a')$, and then execute  state transition.
According to the immediate rewards and state transition, we can obtain the Pareto frontier based on the Pareto optimal solutions.
An example of dynamic update of the dispatching loop based on real-time bicycle rental behavior and state transition is shown in Fig. \ref{fig009}.

As shown in Fig. \ref{fig009}, suppose the capacity of the dispatch truck is 60 bicycles and the initial number of bicycles is 43.
The initial dispatching loop of the truck is $\{(v_0,-9)$, $(v_1,+10)$, $(v_2,-15)$, $(v_5,+12)$, $(v_8,+23)$, $(v_7,+21)$, $(v_4,-18)$, $(v_3,+19)$, $(v_6,-20)\}$.
At time $t_1$, the truck arrives at the parking spot $v_5$, and the number of bicycles to be dispatched is dynamically updated from +18 to +12.
In this case, the number of available bicycles on the truck will be (43-5-10+11-18) = 21, which is lower than the number of bicycles required at spot $v_8$ (28).
Therefore, the truck must changes the dispatching loop and go to the spot $v_4$ to load more bicycles.

The detailed steps of the bicycle dispatching algorithm based on multi-object reinforcement learning (MORL-BD) is described in Algorithm \ref{alg2}.
Since $K$ is the maximal number of search iterations in MORL-BD, $m$ is the number of dispatching loops, and $|A|$ is the number of the action set, so the time complexity of Algorithm 4.1 is $O(Km|A|)$.

\begin{algorithm}[!ht]
\caption{Bicycle dispatching algorithm based on multi-object reinforcement learning (MORL-BD)}
\label{alg2}
\begin{algorithmic}[1]
\REQUIRE ~\\
    $G^{D}_{t+1}$: the bicycle dispatch graph model of $t+1$;\\
    $M$: the dispatch trucks;\\
    $K$: the maximal number of search iterations for MORL-BD;\\
\ENSURE ~\\
    $S(P)$: the Pareto optimal solution of $G^{D}_{t+1}$.
\STATE Generate a set of dispatching loops $L=\{L_{1}, \ldots, L_{m}\}$;
\STATE Initialize a state $s_{0} = [(v_{1}, a_{1}), \ldots, (v_{m}, a_{m})]$ at each dispatching center;
\STATE Initialize a Pareto optimal solution $S(P) = \emptyset$;
\WHILE {$k \leq K$}
\FOR  {each $L_{m}$ in $L$}
\STATE Generate the action set $A_{i}=\{a_{i}\}$ for the current dispatch truck $M_{m}$;
\FOR {each action $a_{i}$ in $A_{i}$}
\STATE Calculate immediate reward vector $\vec{r}(s_{i},a_{i})$ by using Eq. (\ref{eq20});
\STATE Update the state value $\vec{U}(s_{i},a_{i})$ by using Eq. (\ref{eq22});
\ENDFOR
\STATE Find the optimal action $(s', a')$;
\STATE Perform state movement $s \leftarrow s'$;
\ENDFOR
\STATE Form a dispatching solution $\vec{U}$;
\IF {$\vec{U}$ dominates all solutions $\vec{V}$ in $S(P)$}
\STATE Add $\vec{U}$ in $S(P)$: $S(P) \leftarrow \vec{U}$;
\ENDIF
\ENDWHILE
\RETURN $S(P)$.
\end{algorithmic}
\end{algorithm}

\section{Experiments}

\subsection{Experimental Setting}
We conduct extensive comparative experiments on an actual DL-PBS system to compare the proposed MORL-BD algorithm with the multi-objective genetic algorithm (NSGA-II) \cite{deb2002fast}, evolutionary algorithm (MOEA/D) \cite{zhang2007moea}, and particle swarm optimization (MOPSO) algorithm \cite{coello2004handling}.
Large-scale historical bicycle GPS records and cycling trajectory records are gathered from an actual DL-PBS system in China.
There are 2,118,190 GPS records and 372,193,743 trajectory records in Beijing, China.
Firstly, the dataset is divided into 8 spatial subsets according to the districts, and each of them is further split into 94 spatiotemporal subsets by days.
Then, we perform the bicycle parking spot clustering and graph model construction on each spatiotemporal subset, and establish a graph sequence model of bicycle parking spots.
In addition, based on the graph sequence of previous time periods, we forecast the bicycle station layout of the subsequent time periods to create the dispatching demand graph model.
Finally, we perform the comparison algorithms based on the dispatching demand graphs to discuss the experiment results and evaluate the performance of these algorithms.

\subsection{Experiment Result Discussion}
\begin{figure}[!ht]
\centering
\subfigure[Predicted layout of bicycle parking spots]{\includegraphics[width=2.7in]{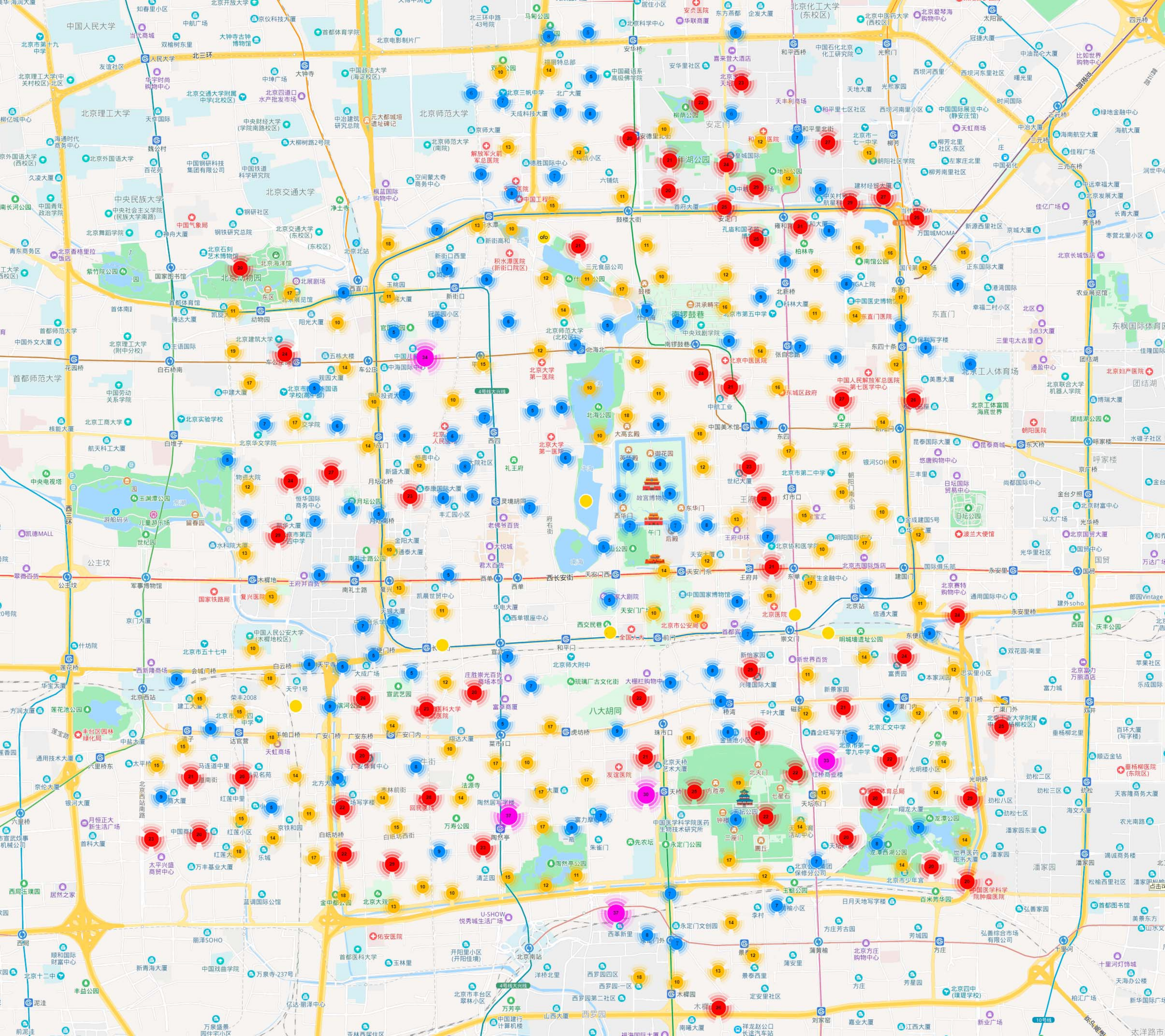}}
 \subfigure[Dispatching demand graph $G^{D}_{t+1}$]{\includegraphics[width=2.7in]{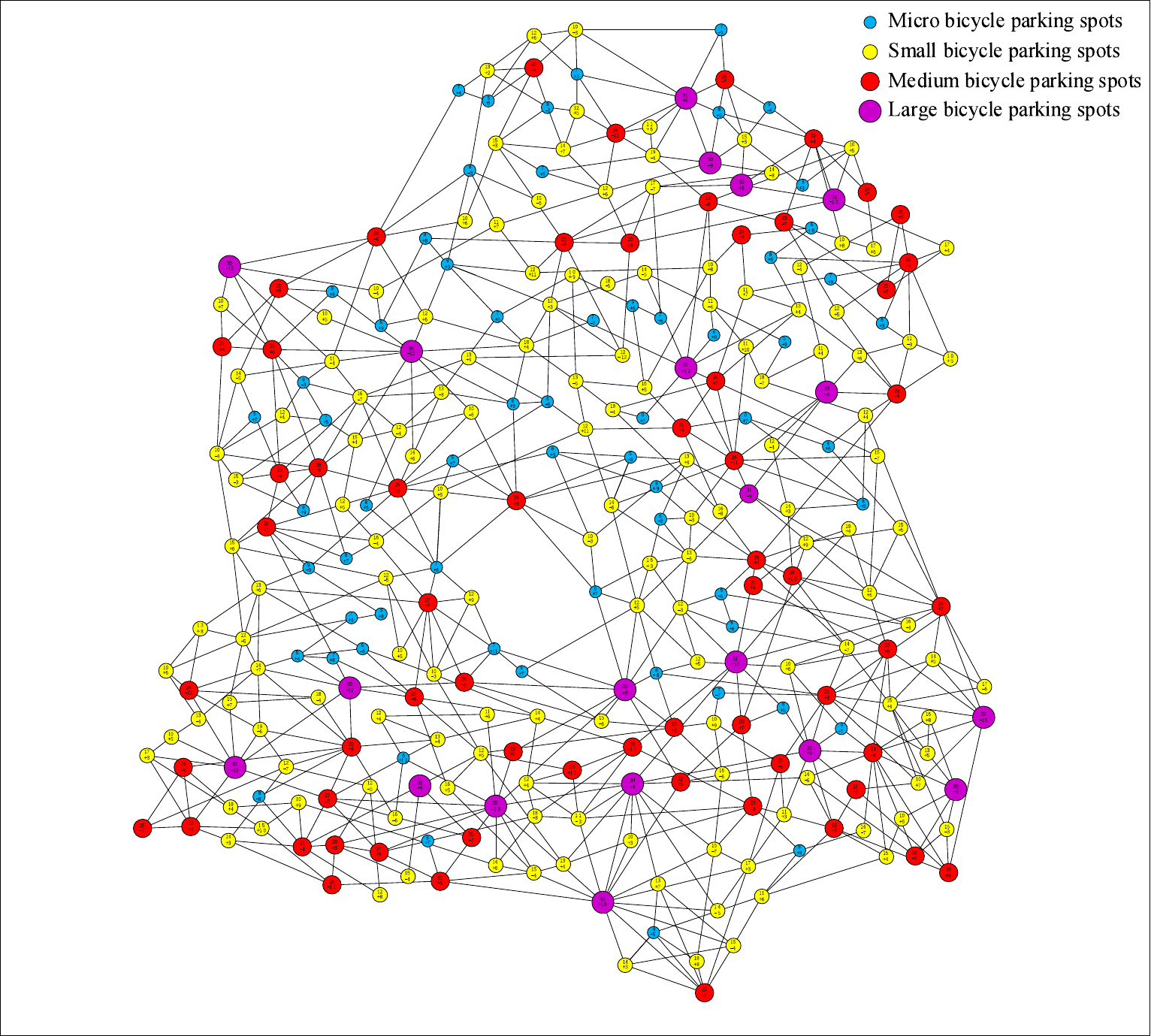}}
 \caption{Predicted layout of bicycle parking spots and the corresponding dispatching demand graph of an actual DL-PBS system in Beijing, China.}
 \label{fig010}
\end{figure}

The actual data of the DL-PBS system in the Dongcheng and Xicheng districts of Beijing is used in this experiment.
The historical records from January 1st, 2019 to October 23, 2019 are gathered for bicycle parking spot clustering and graph model construction.
By using the GGNN algorithm, we predict the layout of bicycle parking spots on October 24, 2019, including the location of each bicycle station and the number of bicycles required at each spot.
The predicted layout of bicycle parking spots and the corresponding dispatching demand graph of an actual DL-PBS system in Beijing is shown in Fig. \ref{fig010}.

As shown in Fig. \ref{fig010}(a), we obtain 56 micro bicycle parking spots (each spot can accommodate 5 to 10 bicycles) and 98 small parking spots (accommodating 10 to 20 bicycles), 61 medium parking spots (accommodating 20 to 30 bicycles) and 20 large parking spots (accommodating more than 30 bicycles).
In addition, based on the actual spot layout on October 23, 2019 and the predicted spot layout on October 24, 2019, we calculate the bicycle dispatching requirements on October 24.
We further construct the corresponding dispatching demand graph model, as shown in Fig. \ref{fig010}(b).

\begin{figure*}[!ht]
\centering
 \subfigure[Dispatching solutions of MORL-BD]{\includegraphics[width=2.7in]{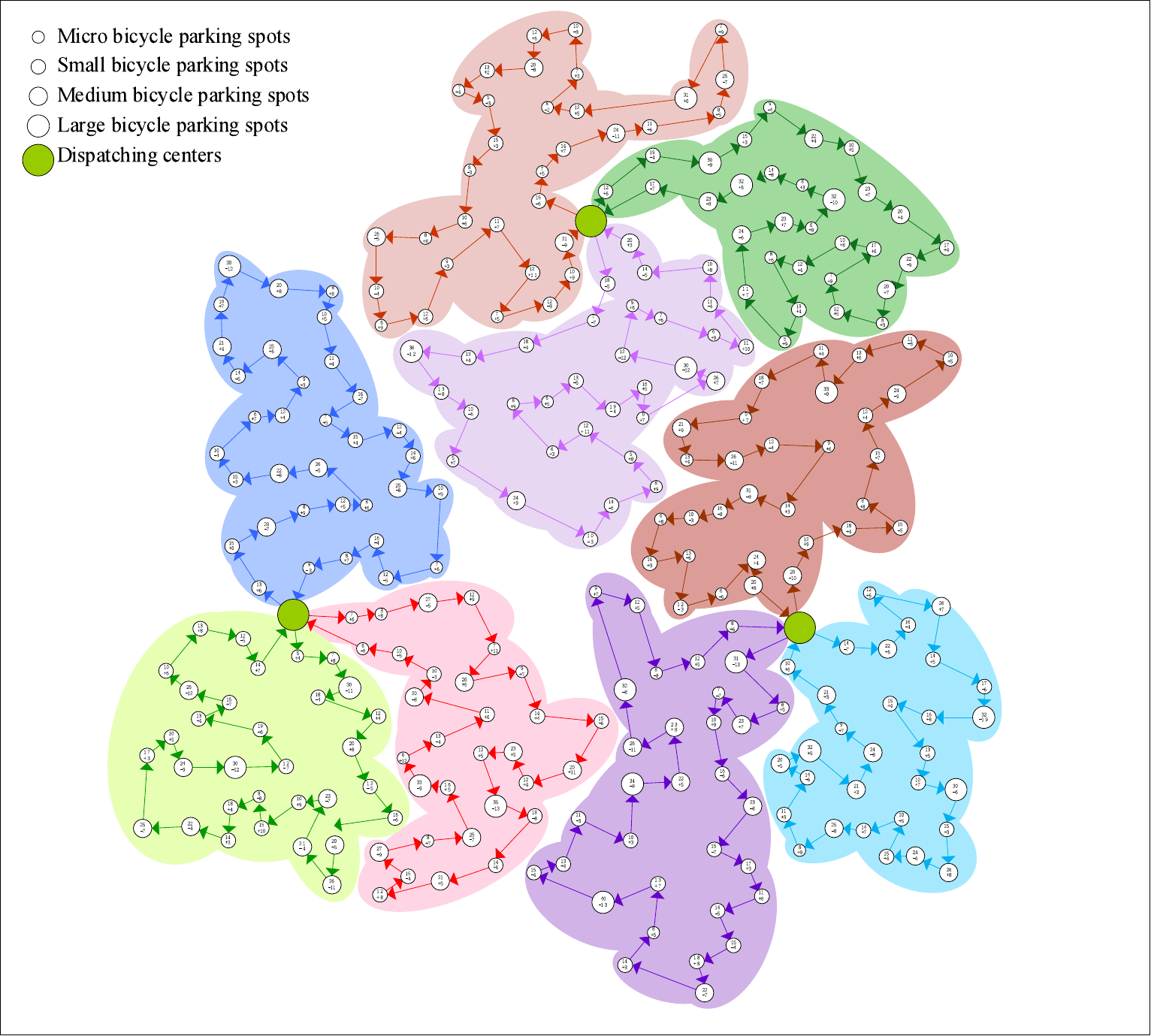}}
 \subfigure[Dispatching solutions of MOPSO]{\includegraphics[width=2.7in]{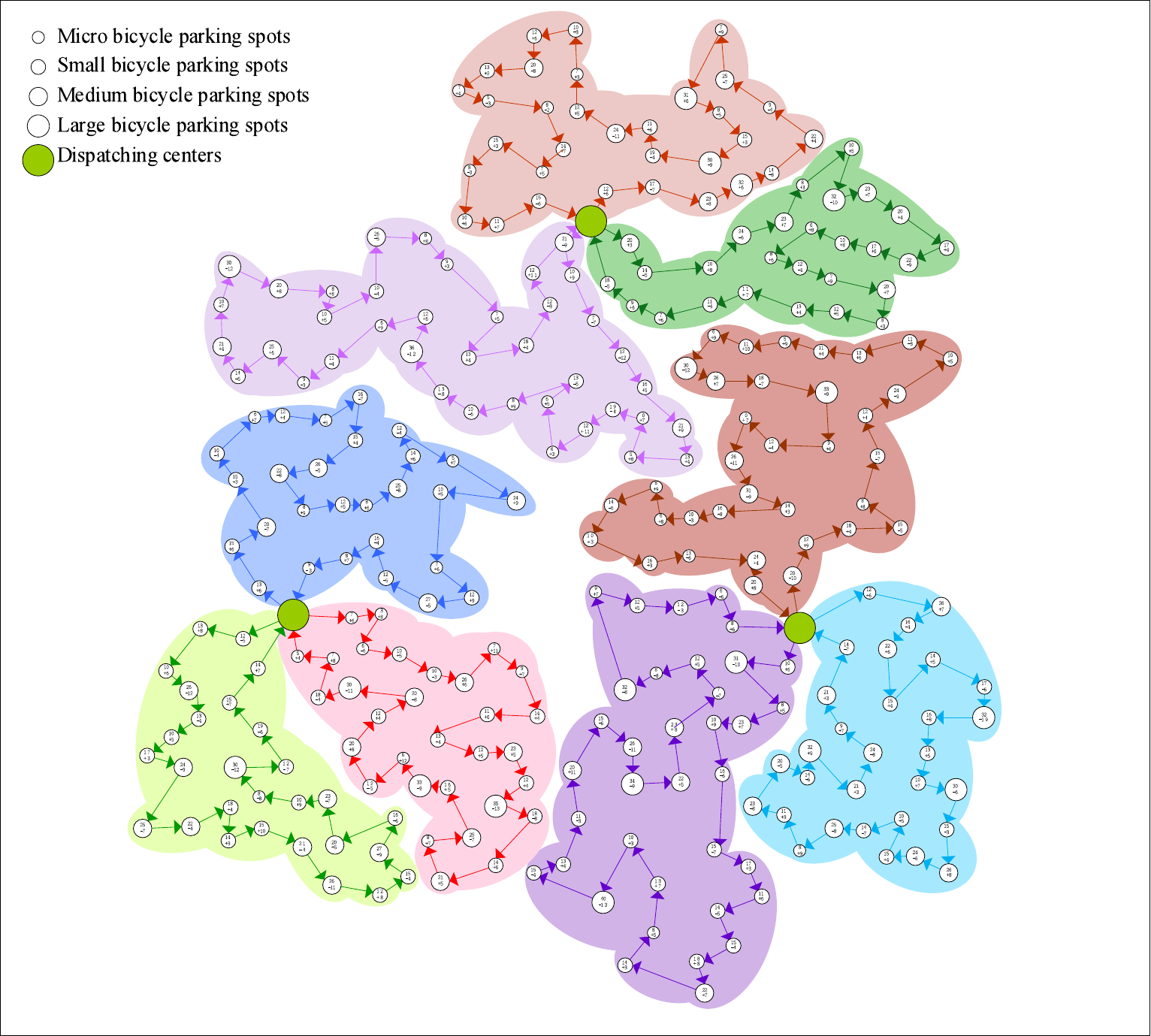}}
 \subfigure[Dispatching solutions of NSGA-II]{\includegraphics[width=2.7in]{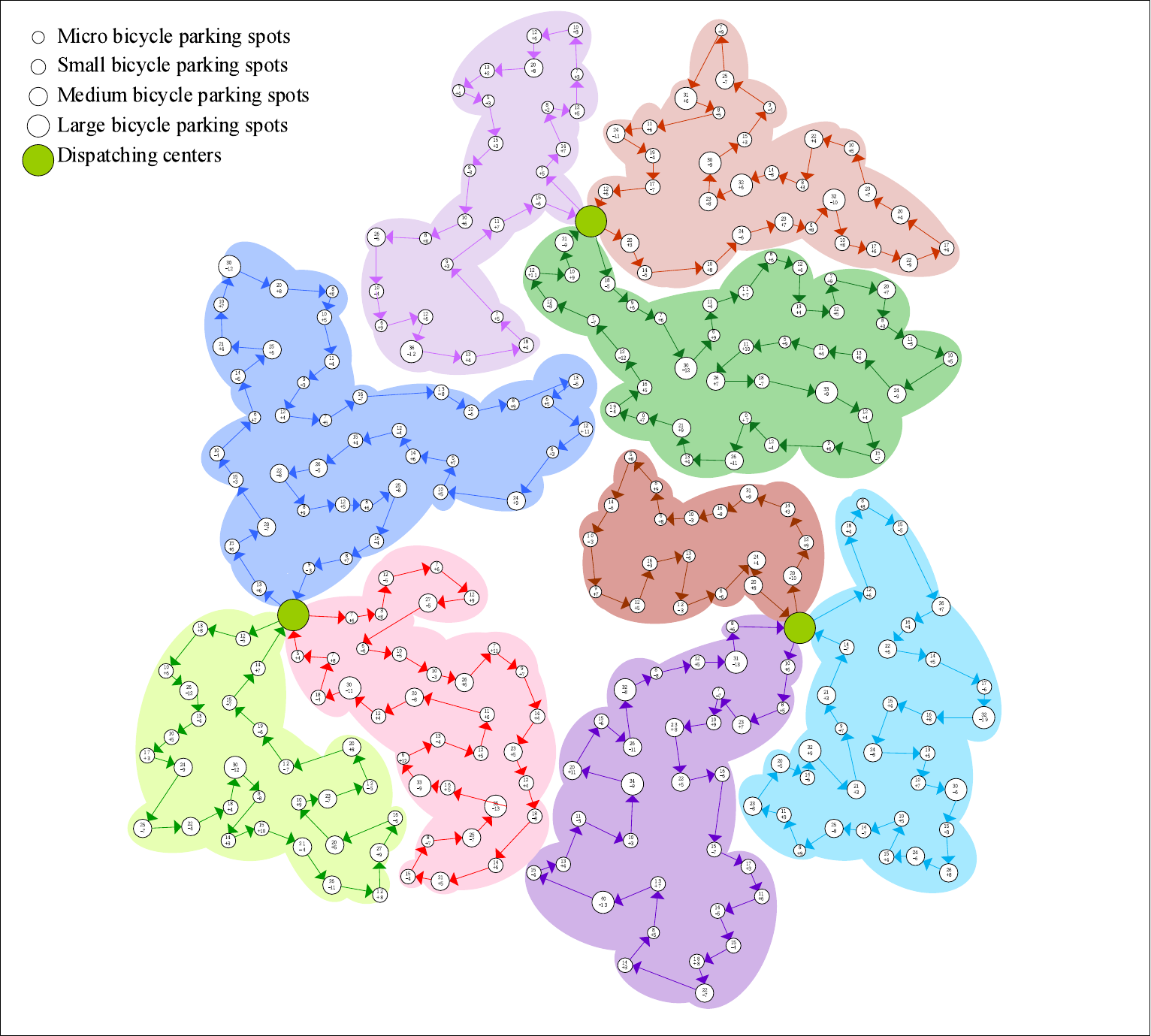}}
 \subfigure[Dispatching solutions of MOEA/D]{\includegraphics[width=2.7in]{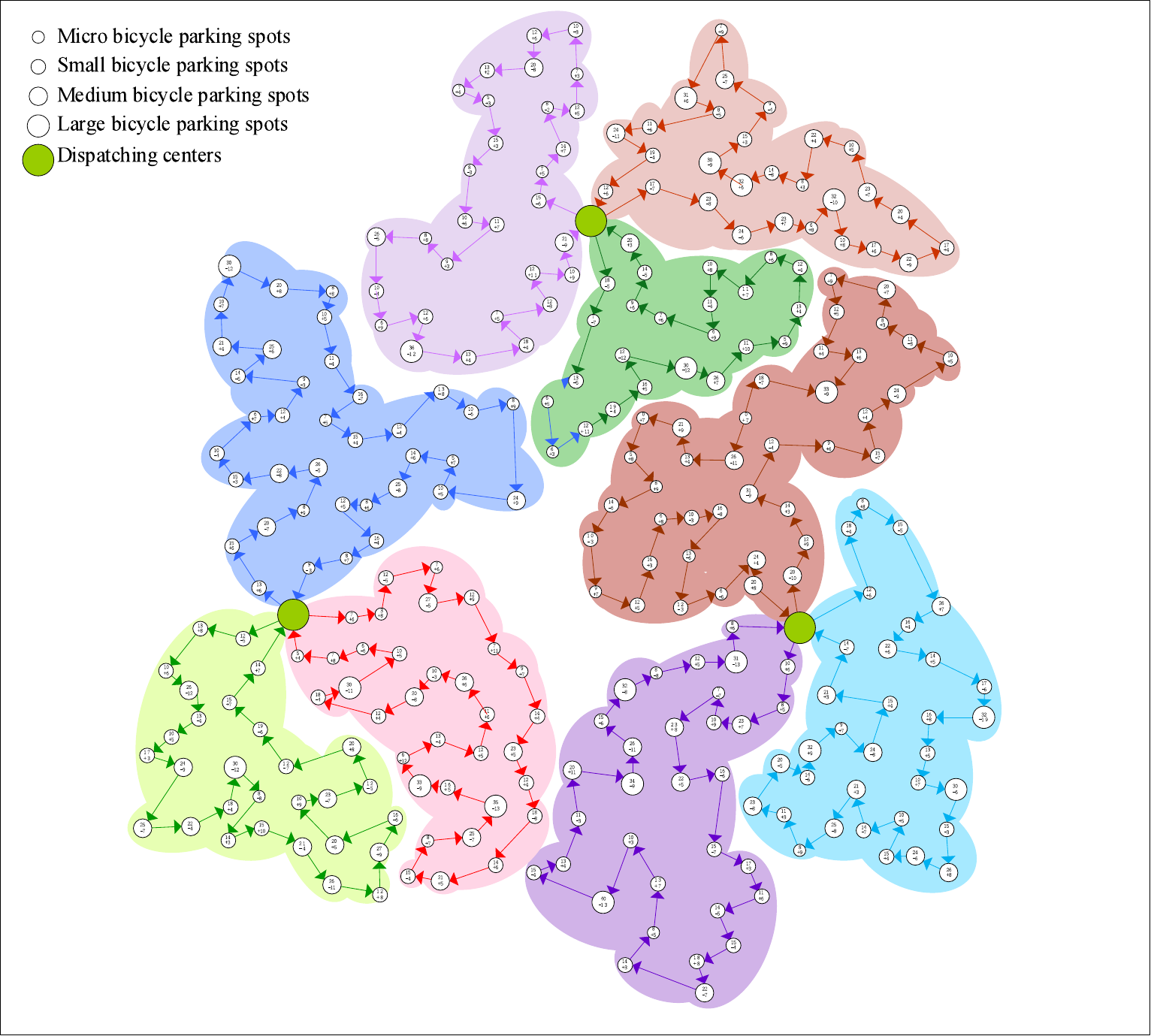}}
 \caption{Dispatching solutions of  MORL-BD, MOPSO,  NSGA-II,  and MOEA/D algorithms.}
 \label{fig011}
\end{figure*}

Given 3 dispatching centers and 9 dispatch trucks, we respectively use the MORL-BD, MOPSO, NSGA-II, and MOEA/D algorithms to find the multi-route bicycle dispatching plans.
The dispatching solutions of all algorithms are shown in Fig. \ref{fig011} and Table \ref{table51}.

\begin{table}[!ht]
\centering
\caption{Value of optimization objectives in dispatching solutions using all methods.}
\label{table51}
\tabcolsep1pt
\begin{tabular}{L{0.6in} C{1.3in} C{1.5in} C{0.6in} C{1.2in}}
\hline
\textbf{Methods} & \textbf{Lengths of dispatch paths (km)} & \textbf{Average number of initial load for each truck} & \textbf{Work balance} & \textbf{Average balance of supply and demand}\\
\hline
MORL-BD & 106.36 & 59.34 & 0.87 & 0.83 \\
MOPSO   & 114.19 & 63.17 & 0.79 & 0.72 \\
NSGA    & 120.87 & 60.21 & 0.78 & 0.80 \\
MOEA/D  & 117.53 & 67.90 & 0.70 & 0.69 \\
\hline
\end{tabular}
\end{table}

As shown in Fig. \ref{fig011}(a), in the solution of MORL-BD, the total length of dispatch paths of all dispatching loops is 106.36 km, the average number of initial load of each dispatch truck is 59.34 (bicycles), the work balance between the dispatch trucks is 0.87, and the average balance of supply and demand at all spots is 0.83.
In contrast, in the solution of MOPSO (shown in Fig. \ref{fig011}(b)), the total length of dispatch paths of all dispatching loops is 114.19 km, the average number of initial load of each dispatch truck is 63.17 (bicycles), the work balance between the dispatch trucks is 0.79, and the average balance of supply and demand at all spots is 0.72.
In the solution of NSGA-II (shown in Fig. \ref{fig011}(c)), the total length of dispatch paths of all dispatching loops is 120.87 km, the average number of initial load of each dispatch truck is 60.21 (bicycles), the work balance between the dispatch trucks is 0.78, and the average balance of supply and demand at all spots is 0.80.
In the solution of MOEA/D (shown in Fig. \ref{fig011}(d)), the total length of dispatch paths of all dispatching loops is 117.53 km, the average number of initial load of each dispatch truck is 67.90 (bicycles), the work balance between the dispatch trucks is 0.70, and the average balance of supply and demand at all spots is 0.69.
Experimental results show that the dispatching solution found by MORL-BD is better than the comparison algorithms in four optimization objectives.
We will evaluate the dispatching costs, Pareto optimality, and algorithm performance of each algorithm in the subsequent sections.

\subsection{Dispatching Cost Comparison}

\begin{figure*}[!ht]
\centering
\subfigure[Different scales of parking spots]{\includegraphics[width=1.73in]{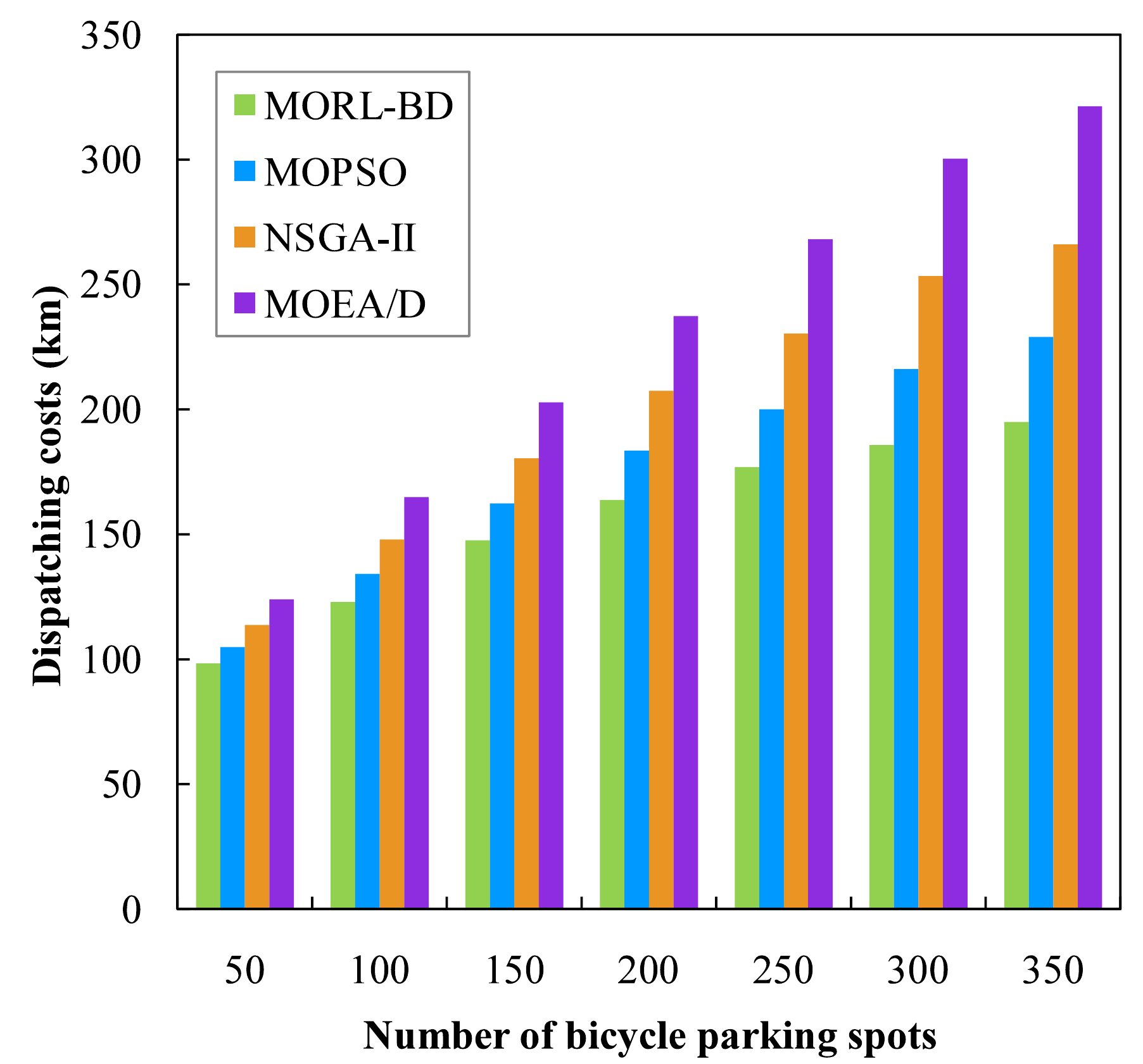}}
 \subfigure[Different amount of bicycles]{\includegraphics[width=1.73in]{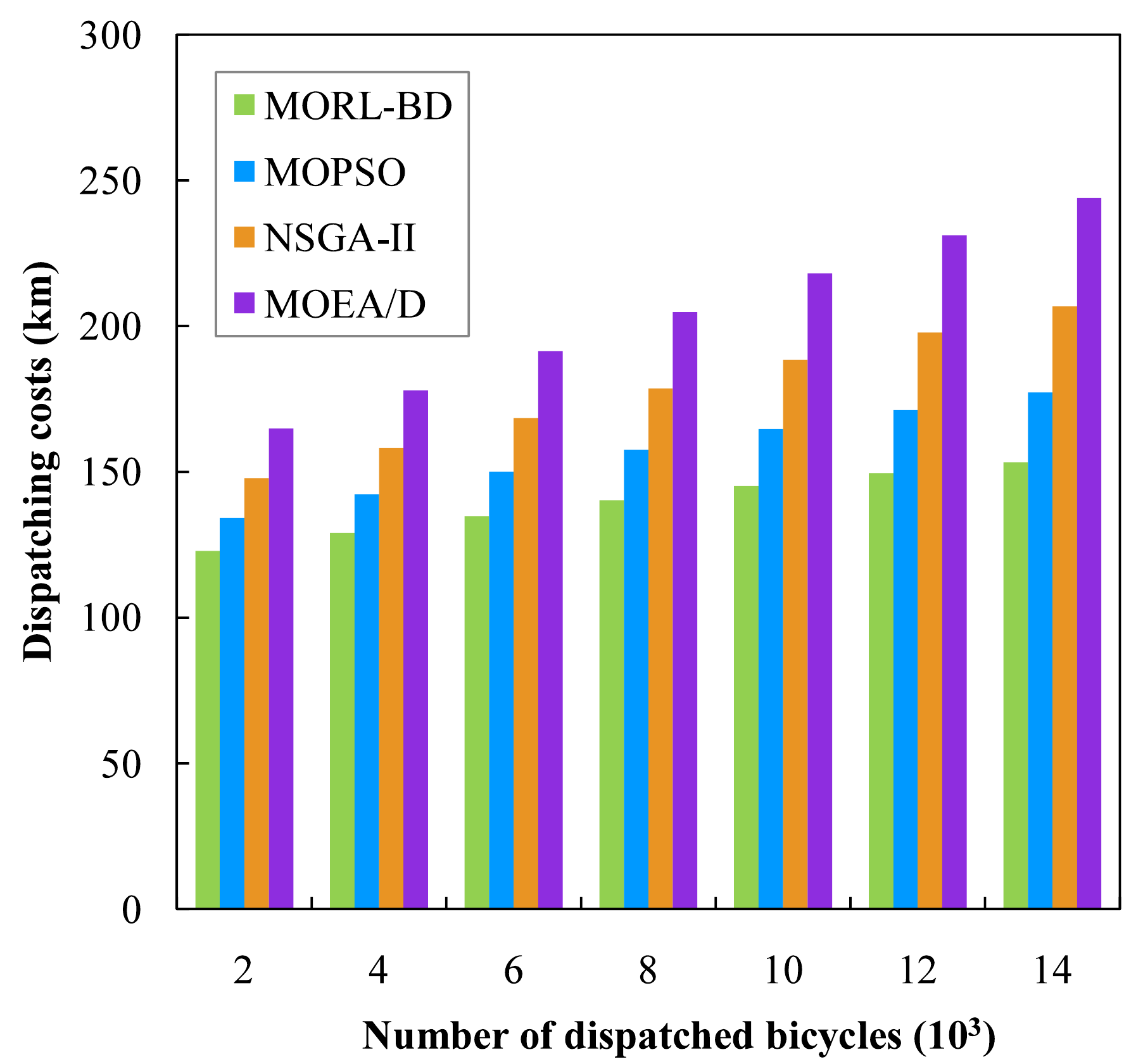}}
 \subfigure[Different amount of dispatch trucks]{\includegraphics[width=1.73in]{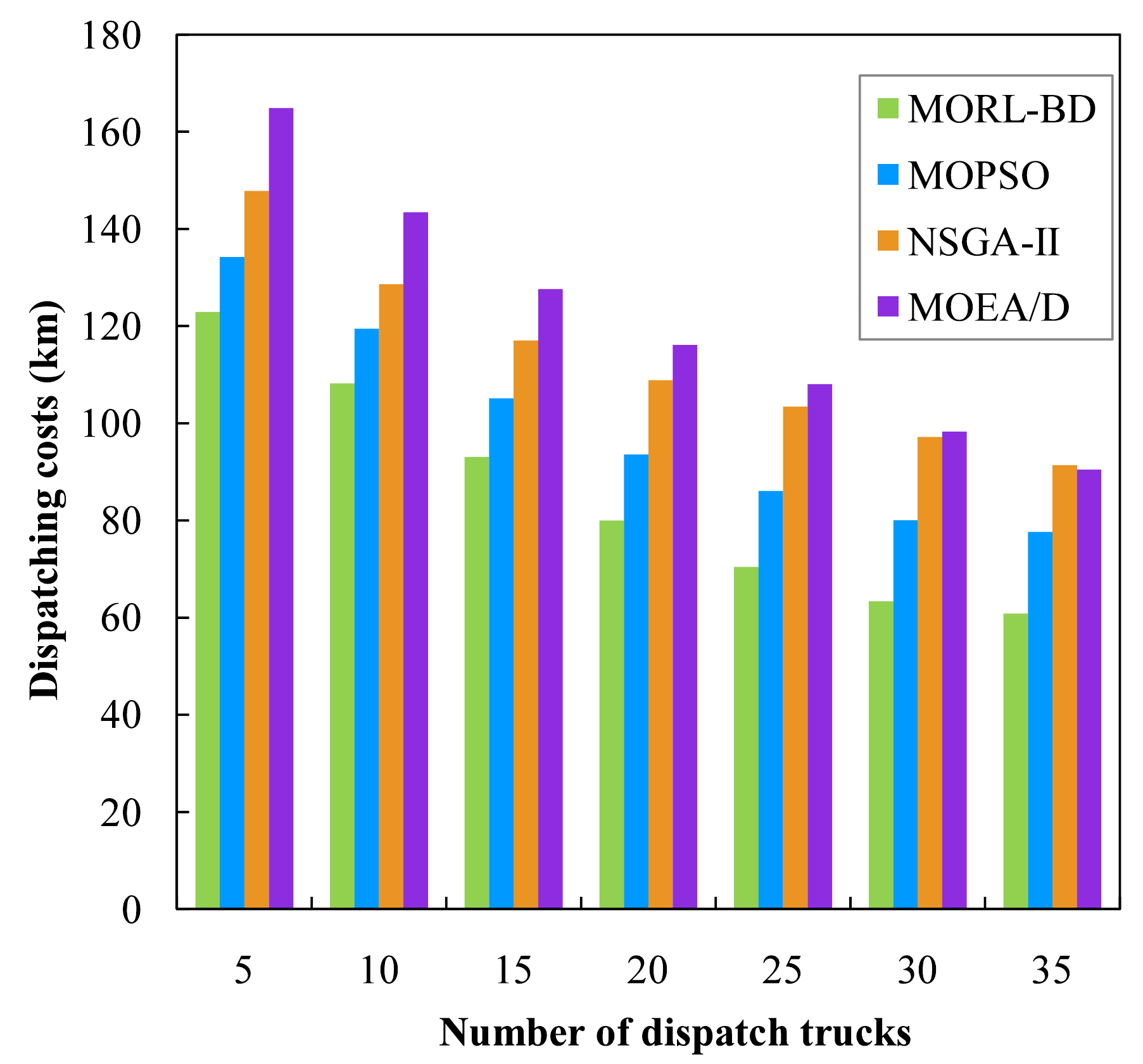}}
 \caption{Dispatching cost comparison of the algorithms on different dispatch cases.}
 \label{fig012}
\end{figure*}

We conduct three groups of comparative experiments to evaluate the dispatching costs of the MORL-BD, MOPSO, NSGA-II, and MOEA/D algorithms.
In the first group of experiments, the number bicycle parking spots is set in the range of [50, 100, 150, 200, 250, 300, 350], with a total of 10,000 dispatched bicycles and 10 dispatch trucks.
In the second group of experiments, we fix the number of bicycle parking spots and dispatch trucks to 100 and 10, respectively, and adjusted the number of dispatched bicycle from $2 \times 10^{3}$ to $14 \times 10^{3}$.
In the third group of experiments, we fix the number of bicycle parking spots and dispatched bicycles to 100 and 10,000, respectively, and adjusted the number of dispatch trucks from 5 to 35.
According to Eq. (\ref{eq04}), we consider the length of dispatch paths as the dispatching cost of each algorithm.
The comparison results of the dispatching costs of these algorithms are shown in Fig. \ref{fig012}.

As can be seen from Fig. \ref{fig012}(a), as the number of bicycle parking spots increases, the length of dispatch paths of each algorithm increases significantly.
However, it is clearly observed that our MORL-BD algorithm has lower dispatching cost than other algorithms.
For example, when the number of bicycle parking spots is equal to 350, the length of dispatch paths of MORL-BD is 195.04 km, the dispatching length of MOPSO is 220.02 km, the dispatching length of NSGA-II is 266.18 km, and the dispatching length of MOEA/D is 321.34 km.
It can be seen from Fig. \ref{fig012}(b) that when a fixed scale of bicycle parking spots, the increase in the number of dispatched bicycles will increase the dispatching costs of these algorithms.
This is because the change in the number of dispatched bicycles at each spot will affect the visiting order of the spots, thereby affecting the dispatching route and its length.
MORL-BD always incurs the lowest cost and is lest sensitive to the increase of number of dispatched bicycles.
It can be seen from Fig. \ref{fig012}(c), with the increase in the number of dispatch trucks, the length of the dispatch paths found by each algorithm gradually becomes shorter and finally tends to be stable.
This is because the increase in the number of dispatch trucks will produce multiple dispatching loops, making each dispatching loop more compact, thereby effectively shortening the length of the dispatch paths.
Again, MORL-BD  incurs the lowest cost.

\subsection{Pareto Frontier and Optimal Vectors}

\begin{figure*}[!ht]
\centering
\subfigure[View of objectives 1 and 2]{\includegraphics[width=1.66in]{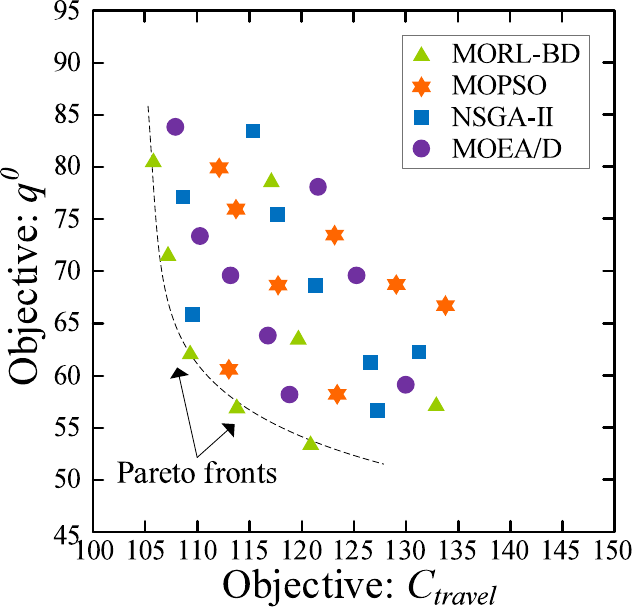}}
 \subfigure[View of objectives 3 and 4]{\includegraphics[width=1.66in]{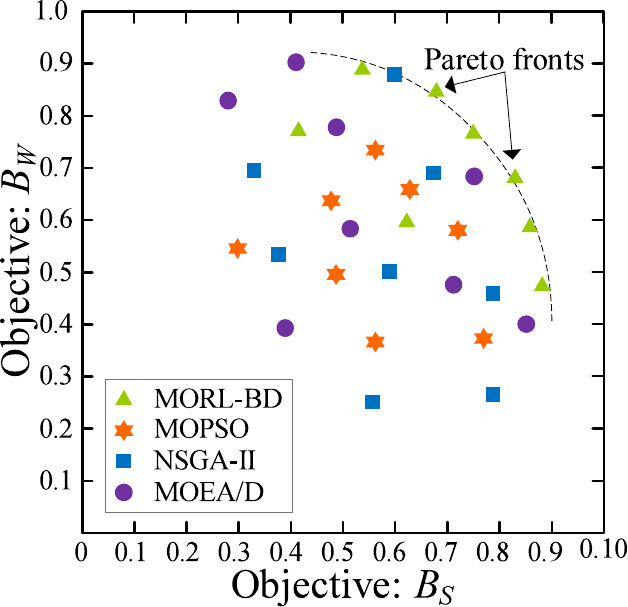}}
 \subfigure[Pareto optimal vectors]{\includegraphics[width=1.86in]{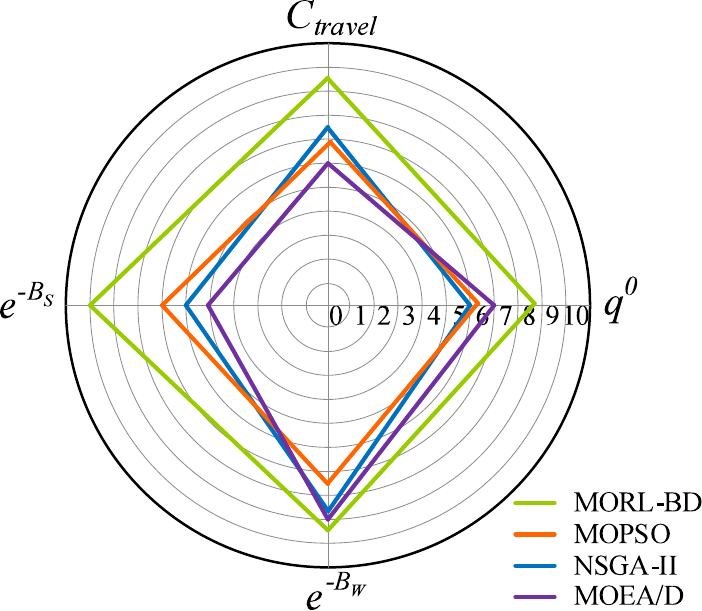}}
 \caption{Pareto frontier and optimal vectors of the algorithms.}
 \label{fig013}
\end{figure*}

We use the first group of experiments in the previous section to discuss the Pareto frontier found by each algorithm, and to evaluate the effectiveness and feasibility of each algorithm.
The experimental results are shown in Fig. \ref{fig013}.

Figs. \ref{fig013}(a), (b), and (c) plot the distribution of Pareto frontier found by the MORL-BD, MOPSO, NSGA-II, and MOEA/D algorithms in the target space of four optimization objectives.
Among them, 74 Pareto solutions found by MORL-BD, while 52 found by MOPSO, 37 found by NSGA-II, and 32 found by MOEA/D.
To clearly show the Pareto frontier of each algorithm, we only draw the top 8 Pareto optimal solutions of each algorithm.
The Pareto optimal objective vector of MORL-BD is distributed in the range of [106.36, 52.08, 0.44, 0.47] $\times$ [135.82, 81.14, 0.88, 0.89],
that of MOPSO is distributed in [114.19, 58.83, 0.35, 0.32] $\times$ [136.25, 80.09, 0.77, 0.72],
that of NSGA-II is distributed in the range of [109.28, 56.11, 0.37, 0.29] $\times$ [132.18, 84.16, 0.81, 0.88],
and that of MOEA/D is distributed in the range of [107.22, 58.54, 0.31, 0.39] $\times$ [131.86, 84.49, 0.85, 0.90].
Hence, the range of the Pareto optimal objective vector found by MORL-BD is wider than the range found by the other algorithms.
In addition, the objective vectors found by MORL-BD dominate the majority of objective vectors found by the other algorithms, which means that MORL-BD can find high-quality Pareto optimal solutions.

\subsection{Algorithm Performance Evaluation}
\begin{figure*}[!ht]
\centering
\subfigure[Different scales of parking spots]{\includegraphics[width=1.73in]{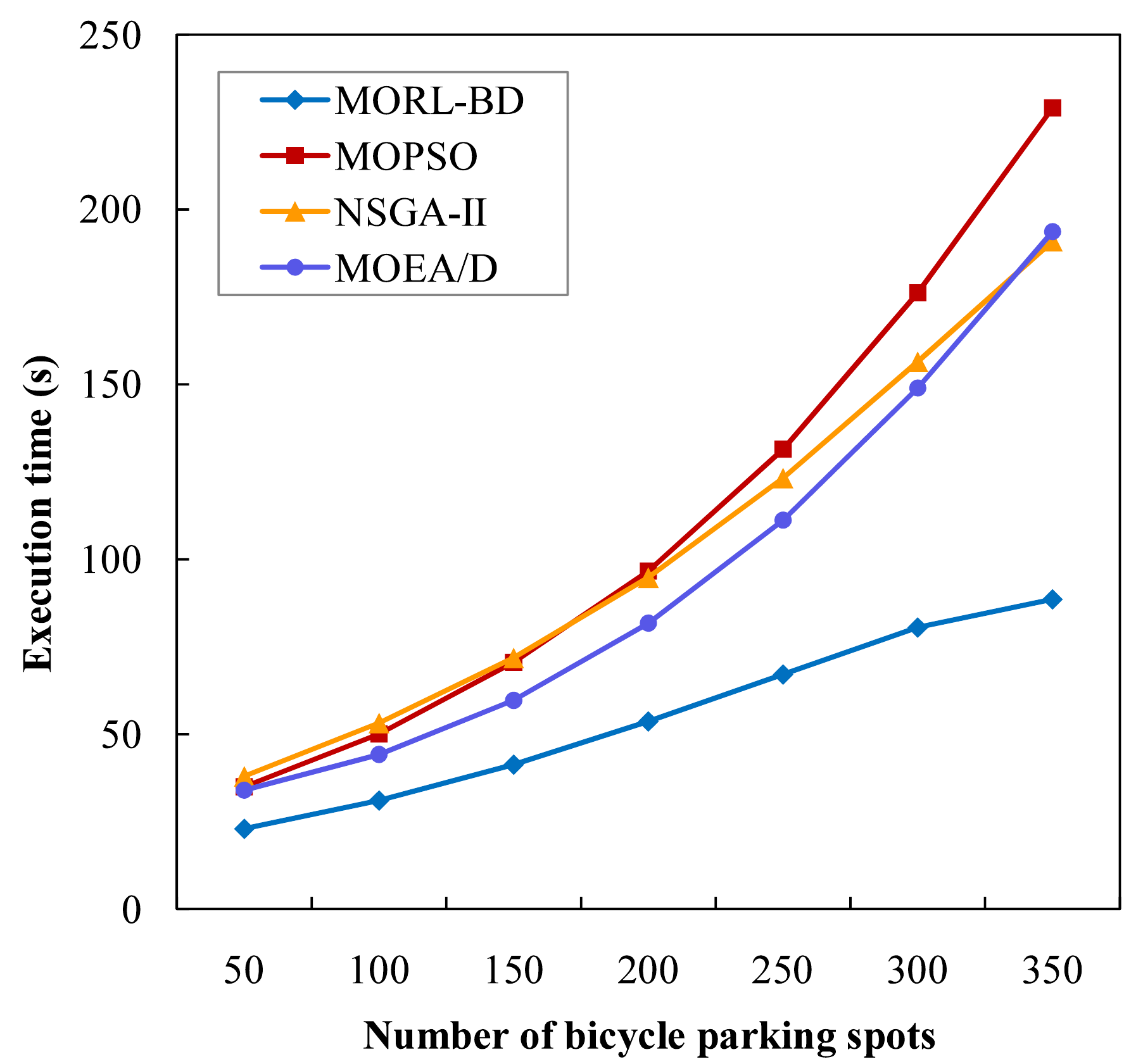}}
 \subfigure[Different number of bicycles]{\includegraphics[width=1.73in]{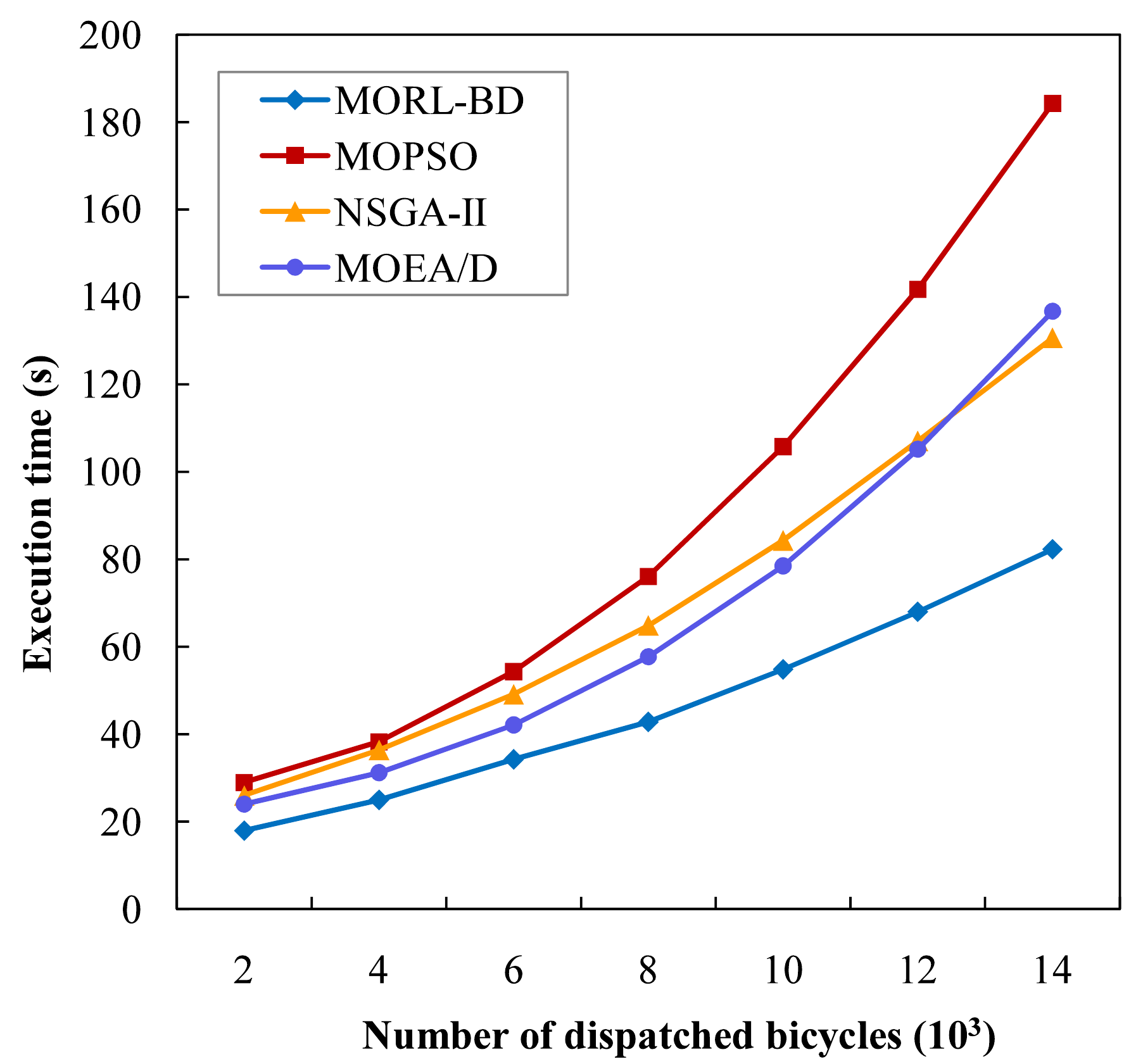}}
 \subfigure[Different number of dispatch trucks]{\includegraphics[width=1.73in]{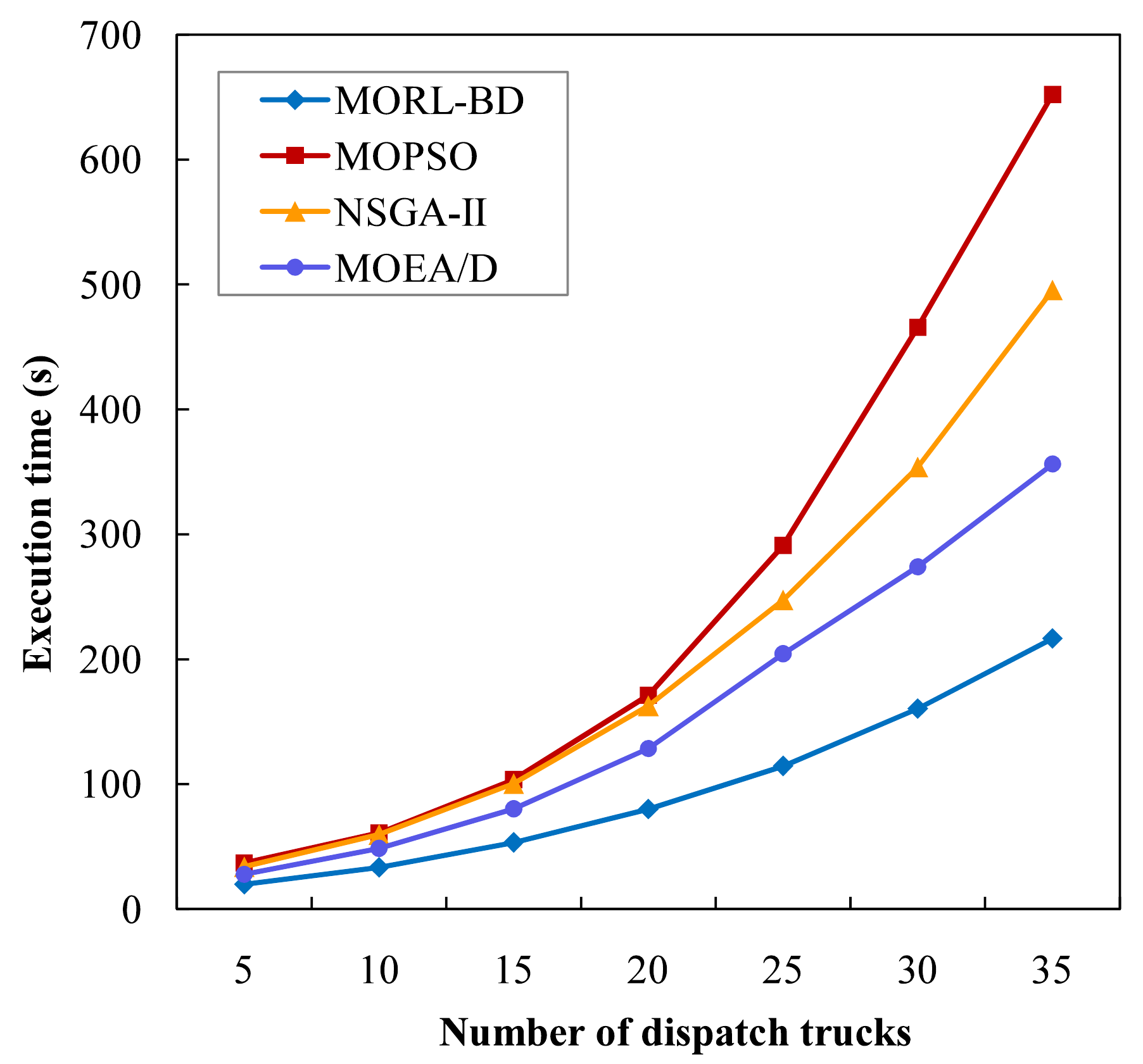}}
 \caption{Performance evaluation of the comparative algorithms under different dispatching situations.}
 \label{fig014}
\end{figure*}

We conduct three groups of comparative experiments to evaluate the performance of MORL-BD, MOPSO, NSGA-II, and MOEA/D algorithms.
The first group is carry out under parking spots of different scales, with the number of parking spots gradually increasing from 50 to 350.
The second group is conducted on dispatched bicycles of different scales, where the number of bicycles gradually increases from 2,000 to 14,000.
The third group is conducted under a different number of dispatch trucks, increasing from 5 to 35.
In each comparison experiment, all algorithms are executed on a high-performance computer equipped with Intel Core i5-6400 8-core CPU, 32 GB DRAM, and 2 TB main memory.
The performance comparison results of these algorithms are shown in Fig. \ref{fig014}.

It can be obvious from Fig. \ref{fig014} that the execution time of the comparison algorithms increases with the increase of the data scale.
However, in each case, the execution time of our MORL-BD algorithm is significantly shorter than that of the comparison algorithms.
For example, when the number of bicycle parking spots is equal to 200, the execution time of MORL-BD is 53.68 (s), while the execution times of MOEA/D, NSGA-II and MOPSO are 96.74 (s), 94.80(s) and 81.76(s), respectively, as shown in Fig. \ref{fig014}(a).
In addition, the increase in the number of dispatch trucks significantly affects the number of iterations in each algorithm, thereby resulting in an increase in the execution time of each algorithm.
For example, when the number of dispatch trucks increases from 5 to 35, the execution time of MORL-BD increases from 21.08 (s) to 216.64 (s), while the execution time of MOEA/D increases from 28.34 (s) to 357.25 (s), the execution time of NSGA-II increases from 34.98 (s) to 495.38 (s), and the execution time of MOPSO increases from 37.91 (s) to 652.10 (s), as shown in Fig. \ref{fig014}(c).
Similar observation can be made when the number of dispatched bicycles is increased as shown in Fig. \ref{fig014}(b).
Therefore, experimental results show that MORL-BD is superior to the comparison algorithms in performance, and can efficiently achieve the optimization objectives of bicycle dispatching.

\section{Conclusions}

This paper presented the MORL-based bicycle dispatching (MORL-BD) algorithm, and provided the optimal bicycle dispatching solutions for the DL-PBS system.
The requirements of multi-route bicycle dispatching of DL-PBS systems were described and a dispatching demand graph model was created.
The multi-route bicycle dispatching problem was defined as a multi-objective optimization problem with four optimization objectives.
In addition, the MORL-BD algorithm was used to search candidate Pareto optimal solutions for bicycle dispatching routes.
The Pareto optimal solution found in each action was saved to find out the Pareto frontier.
Experimental results on the actual DL-PBS datasets show that the proposed MORL-BD algorithm outperforms the comparison algorithms in terms of effectiveness and performance.

\section*{Acknowledgment}
This work is partially funded by the National Key R\&D Program of China (Grant No. 2020YFB2104000),
the National Outstanding Youth Science Program of National Natural Science Foundation of China (Grant No. 61625202),
the Program of National Natural Science Foundation of China (Grant No. 61751204),
the International (Regional) Cooperation and Exchange Program of National Natural Science Foundation of China (Grant No. 61860206011),
the Natural Science Foundation of Hunan Province (Grant No. 2020JJ5084),
and the International Postdoctoral Exchange Fellowship Program (Grant No. 20180024).
This work is also supported in part by NSF under grants III-1763325, III-1909323, and SaTC-1930941.

\bibliographystyle{abbrv}
\bibliography{sample-base}

\begin{thebibliography}{10}

\bibitem{ben1980characterization}
A.~Ben-Tal.
\newblock Characterization of pareto and lexicographic optimal solutions.
\newblock In {\em Multiple criteria decision making theory and application},
  pages 1--11. Springer, Springer, 1980.

\bibitem{chen2020tii}
J.~Chen, K.~Li, K.~Li, P.~S. Yu, and Z.~Zeng.
\newblock Dynamic planning bicycle stations for dock-free public
  bicycle-sharing networks based on gated graph neural networks.
\newblock {\em ACM Trans. Intell. Syst. Technol.}, 99(1):1, 2021.

\bibitem{chen2018personalized}
Y.~Chen, W.~Yan, C.~Li, Y.~Huang, and L.~Yang.
\newblock Personalized optimal bicycle trip planning based on q-learning
  algorithm.
\newblock In {\em WCNC'18}, pages 1--6. IEEE, IEEE, 2018.

\bibitem{coello2004handling}
C.~Coello, G.~Pulido, and S.~Lechuga.
\newblock Handling multiple objectives with particle swarm optimization.
\newblock {\em IEEE Trans. Evol. Comput.}, 8(3):256--279, 2004.

\bibitem{deb2002fast}
K.~Deb, A.~Pratap, S.~Agarwal, and T.~Meyarivan.
\newblock A fast and elitist multiobjective genetic algorithm: Nsga-ii.
\newblock {\em IEEE Trans. Evol. Comput.}, 6(2):182--197, 2002.

\bibitem{duan2019optimizing}
Y.~Duan and J.~Wu.
\newblock Optimizing rebalance scheme for dock-less bike sharing systems with
  adaptive user incentive.
\newblock In {\em MDM'19}, pages 176--181. IEEE, 2019.

\bibitem{ghosh2017dynamic}
S.~Ghosh, P.~Varakantham, Y.~Adulyasak, and P.~Jaillet.
\newblock Dynamic repositioning to reduce lost demand in bike sharing systems.
\newblock {\em J. Artif. Intell. Res.}, 58:387--430, 2017.

\bibitem{hu2012data}
J.~Hu, C.~J. Xue, Q.~Zhuge, W.-C. Tseng, and E.~H.-M. Sha.
\newblock Data allocation optimization for hybrid scratch pad memory with sram
  and nonvolatile memory.
\newblock {\em IEEE Trans. Very Large Scale Integr. (VLSI) Syst.},
  21(6):1094--1102, 2012.

\bibitem{hu2014optimal}
S.-R. Hu and C.-T. Liu.
\newblock An optimal location model for a bicycle sharing program with truck
  dispatching consideration.
\newblock In {\em ITSC'14}, pages 1775--1780. IEEE, 2014.

\bibitem{gnn08}
Y.~Li, D.~Tarlow, M.~Brockschmidt, and R.~Zemel.
\newblock Gated graph sequence neural networks.
\newblock {\em arXiv e-prints}, June 2017.

\bibitem{li2019citywide}
Y.~Li and Y.~Zheng.
\newblock Citywide bike usage prediction in a bike-sharing system.
\newblock {\em IEEE Trans. Knowledge Data Eng.}, 32(6):1079--1091, 2019.

\bibitem{li2018dynamic}
Y.~Li, Y.~Zheng, and Q.~Yang.
\newblock Dynamic bike reposition: A spatio-temporal reinforcement learning
  approach.
\newblock In {\em SIGKDD'18}, pages 1724--1733. ACM, 2018.

\bibitem{liu2018vehicle}
D.~Liu, H.~Dong, T.~Li, J.~Corcoran, and S.~Ji.
\newblock Vehicle scheduling approach and its practice to optimise public
  bicycle redistribution in hangzhou.
\newblock {\em IET Intell. Transp. Syst.}, 12(8):976--985, 2018.

\bibitem{ma2019data}
M.~Ma, S.~M. Preum, M.~Y. Ahmed, W.~T{\"a}rneberg, A.~Hendawi, and J.~A.
  Stankovic.
\newblock Data sets, modeling, and decision making in smart cities: A survey.
\newblock {\em ACM Transactions on Cyber-Physical Systems}, 4(2):1--28, 2019.

\bibitem{mao2019novel}
D.~Mao, Z.~Hao, Y.~Wang, and S.~Fu.
\newblock A novel dynamic dispatching method for bicycle-sharing system.
\newblock {\em ISPRS Int. Geo-Inf.}, 8(3):117, 2019.

\bibitem{mimura2019bike}
T.~Mimura, S.~Ishiguro, S.~Kawasaki, and Y.~Fukazawa.
\newblock Bike-share demand prediction using attention based sequence to
  sequence and conditional variational autoencoder.
\newblock In {\em PredictGIS'19}, pages 41--44. ACM, 2019.

\bibitem{pan2019deep}
L.~Pan, Q.~Cai, Z.~Fang, P.~Tang, and L.~Huang.
\newblock A deep reinforcement learning framework for rebalancing dockless bike
  sharing systems.
\newblock In {\em AAAI'19}, volume~33, pages 1393--1400, 2019.

\bibitem{parisi2016multi}
S.~Parisi, M.~Pirotta, and M.~Restelli.
\newblock Multi-objective reinforcement learning through continuous pareto
  manifold approximation.
\newblock {\em J. Artif. Intell. Res.}, 57:187--227, 2016.

\bibitem{ren2020rebalancing}
C.~Ren, L.~An, Z.~Gu, Y.~Wang, and Y.~Gao.
\newblock Rebalancing the car-sharing system with reinforcement learning.
\newblock {\em World Wide Web}, pages 1--21, 2020.

\bibitem{ruiz2017temporal}
M.~Ruiz-Montiel, L.~Mandow, and J.-L. P{\'e}rez-de-la Cruz.
\newblock A temporal difference method for multi-objective reinforcement
  learning.
\newblock {\em Neurocomputing}, 263:15--25, 2017.

\bibitem{sun2018sharing}
Y.~Sun.
\newblock Sharing and riding: how the dockless bike sharing scheme in china
  shapes the city.
\newblock {\em Urban Science}, 2(3):68, 2018.

\bibitem{tang2018bikeshare}
G.~Tang, S.~Keshav, L.~Golab, and K.~Wu.
\newblock Bikeshare pool sizing for bike-and-ride multimodal transit.
\newblock {\em IEEE Trans. Intell. Transp. Syst.}, 19(7):2279--2289, 2018.

\bibitem{wu2020challenges}
J.~Wu.
\newblock Challenges and opportunities in algorithmic solutions for
  re-balancing in bike sharing systems.
\newblock {\em Tsinghua Sci. Technol.}, 25(6):721--733, 2020.

\bibitem{yan2019top}
L.~Yan and H.~Shen.
\newblock Top: Optimizing vehicle driving speed with vehicle trajectories for
  travel time minimization and road congestion avoidance.
\newblock {\em ACM Transactions on Cyber-Physical Systems}, 4(2):1--25, 2019.

\bibitem{yang2019estimating}
F.~Yang, F.~Ding, X.~Qu, and B.~Ran.
\newblock Estimating urban shared-bike trips with location-based social
  networking data.
\newblock {\em Sustainability}, 11(11):3220, 2019.

\bibitem{yang2019generalized}
R.~Yang, X.~Sun, and K.~Narasimhan.
\newblock A generalized algorithm for multi-objective reinforcement learning
  and policy adaptation.
\newblock In {\em Advances in Neural Information Processing Systems}, pages
  14636--14647, 2019.

\bibitem{yang2019mobility}
Z.~Yang, J.~Chen, J.~Hu, Y.~Shu, and P.~Cheng.
\newblock Mobility modeling and data-driven closed-loop prediction in
  bike-sharing systems.
\newblock {\em IEEE Trans. Intell. Transport. Syst.}, 20(12):4488--4499, 2019.

\bibitem{yi2019rebalancing}
P.~Yi, F.~Huang, and J.~Peng.
\newblock A rebalancing strategy for the imbalance problem in bike-sharing
  systems.
\newblock {\em Energies}, 12(13):2578, 2019.

\bibitem{yoshida2019practical}
A.~Yoshida, Y.~Yatsushiro, N.~Hata, T.~Higurashi, N.~Tateiwa, T.~Wakamatsu,
  A.~Tanaka, K.~Nagamatsu, and K.~Fujisawa.
\newblock Practical end-to-end repositioning algorithm for managing
  bike-sharing system.
\newblock In {\em BigData'19}, pages 1251--1258. IEEE, 2019.

\bibitem{zhang2016last}
D.~Zhang, J.~Zhao, F.~Zhang, R.~Jiang, T.~He, and N.~Papanikolopoulos.
\newblock Last-mile transit service with urban infrastructure data.
\newblock {\em ACM Transactions on Cyber-Physical Systems}, 1(2):1--26, 2016.

\bibitem{zhang2007moea}
Q.~Zhang and H.~Li.
\newblock Moea/d: A multiobjective evolutionary algorithm based on
  decomposition.
\newblock {\em IEEE Trans. Evol. Comput.}, 11(6):712--731, 2007.

\bibitem{zhao2019study}
J.~Zhao, Y.~Li, H.~Jia, H.~Jian, and J.~Cai.
\newblock Study on allocation scheme of bicycle sharing without piles.
\newblock In {\em CICTP'19}, pages 1603--1614. 2019.

\end{thebibliography}

\end{document}